\documentclass[sigconf]{acmart}

\usepackage{multirow}
\usepackage{multicol}
\usepackage{tabularx}
\usepackage{pifont}
\usepackage{makecell}

\AtBeginDocument{%
  }

\setcopyright{acmlicensed}
\copyrightyear{2025}
\acmYear{2025}
\acmDOI{10.1145/3746027.3755556}
\acmConference[MM '25] {Proceedings of the 33rd ACM International Conference on Multimedia}{October 27--31, 2025}{Dublin, Ireland.}
\acmBooktitle{Proceedings of the 33rd ACM International Conference on Multimedia (MM '25), October 27--31, 2025, Dublin, Ireland}
\acmISBN{979-8-4007-2035-2/2025/10}

\acmSubmissionID{mfp4433}

\begin{document}

\title{EventFormer: A Node-graph Hierarchical Attention Transformer for Action-centric Video Event Prediction}

\author{Qile Su}
\authornote{Both authors contributed equally to this research.}
\orcid{0000-0002-7761-3636}
\affiliation{%
  %\department{National Superior College for Engineers}
  \institution{Beihang University}
  % \institution{State Key Laboratory of Virtual Reality Technology and Systems, Beihang University}
  \city{Beijing}
  \country{China}
}
\email{zy2423342@buaa.edu.cn}

\author{Shoutai Zhu}
\authornotemark[1]
\orcid{0009-0009-6800-0908}
\affiliation{%
  %\department{School of Computer Science and Engineering}
  \institution{Beihang University}
  % \institution{State Key Laboratory of Virtual Reality Technology and Systems, Beihang University}
  \city{Beijing}
  \country{China}
}
\email{shoutaizhu@buaa.edu.cn}

\author{Shuai Zhang}
\orcid{0009-0001-8549-3179}
\affiliation{%
  %\department{School of Intelligence Science and Technology}
  \institution{University of Science and Technology Beijing}
  \city{Beijing}
  \country{China}
}
\email{u202140073@xs.ustb.edu.cn}

\author{Baoyu Liang}
\orcid{0000-0002-2365-589X}
\affiliation{%
  %\department{School of Computer Science and Engineering}
  \institution{Beihang University}
  %\institution{State Key Laboratory of Virtual Reality Technology and Systems, Beihang University}
  \city{Beijing}
  \country{China}
}
\email{liangbaoyu96@buaa.edu.cn}

\author{Chao Tong}
\authornote{Corresponding author}
\orcid{0000-0003-4414-4965}
\affiliation{%
  \department{School of Computer Science and Engineering}
  \department{State Key Laboratory of Virtual Reality Technology and Systems}
  \institution{Beihang University}
  \city{Beijing}
  \country{China}
}
\email{tongchao@buaa.edu.cn}

\renewcommand{\shortauthors}{Qile Su, Shoutai Zhu, Shuai Zhang, Baoyu Liang, and Chao Tong}

\begin{abstract}
Script event induction, which aims to predict the subsequent event based on the context, is a challenging task in NLP, achieving remarkable success in practical applications. However, human events are mostly recorded and presented in the form of videos rather than scripts, yet there is a lack of related research in the realm of vision. To address this problem, we introduce AVEP (Action-centric Video Event Prediction), a task that distinguishes itself from existing video prediction tasks through its incorporation of more complex logic and richer semantic information. We present a large structured dataset, which consists of about $35K$ annotated videos and more than $178K$ video clips of event, built upon existing video event datasets to support this task. The dataset offers more fine-grained annotations, where the atomic unit is represented as a multimodal event argument node, providing better structured representations of video events. Due to the complexity of event structures, traditional visual models that take patches or frames as input are not well-suited for AVEP. We propose EventFormer, a node-graph hierarchical attention based video event prediction model, which can capture both the relationships between events and their arguments and the coreferencial relationships between arguments. We conducted experiments using several SOTA video prediction models as well as LVLMs on AVEP, demonstrating both the complexity of the task and the value of the dataset. Our approach outperforms all these video prediction models. We will release the dataset and code for replicating the experiments and annotations.

\end{abstract}


\begin{teaserfigure}
  \includegraphics[width=\textwidth]{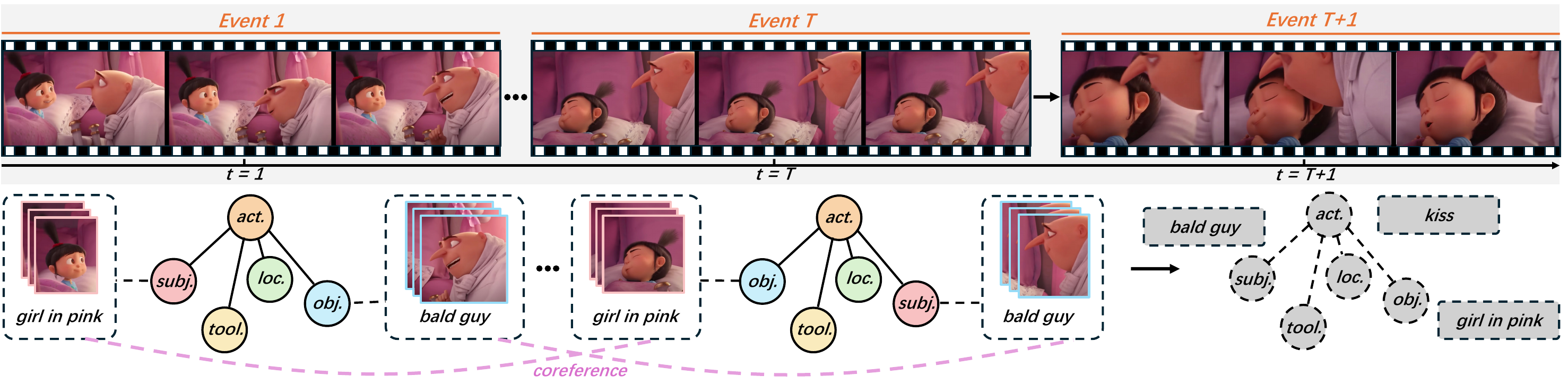}
  \caption{A video event graph chain $\mathcal{C}$ is a series of observed historical video events, represented as video event graphs, where each node in the graph contains both visual and textual representations, and the edges denote argument roles. Given a $\mathcal{C}$, the proposed AVEP task requires the model to predict the verb of future events along with their corresponding arguments.}
  \Description{}
  \label{fig:top}
\end{teaserfigure}

% \received{11 April 2025}
% \received[accepted]{6 July 2025}

\maketitle

\section{Introduction}
\label{sec:intro}

Event prediction has been a key research problem in the field of artificial intelligence \cite{zhao2021event} for a long time, which can not only be used for security surveillance to anticipate potentially harmful behaviors \cite{rumi2018crime} but also play a significant role in embodied intelligence \cite{persia2020fast}. The task of event prediction can trace back to Granroth-Wilding and Clark \cite{granroth2016happens}, in which it was first formally defined as script event induction.

%Given a series of events extracted from a script, script event induction requires the model to predict the possible future event. Since the task was introduced, many scholars have conducted excellent research to tackle this challenge, showing promising results in practical deployment.
% However, due to historical dependencies, the model must be capable of analyzing significant events from past occurrences and also possess strong logical reasoning abilities, making this a challenging area for further research.

It is worth noting that in real life and practical applications, human events are often recorded in multimodal forms. However, there is a lack of research on event prediction in the realm of vision. Several similar video prediction tasks have been proposed, such as action anticipation and visual reasoning, while video event prediction differs from these tasks in the following ways:
\textbf{Richer semantic information} \cite{Liang2025videvent, wang2025fosteringvideoreasoningnextevent}. An event typically comprises subject, action, object, location, and relationship between arguments, with action representing a higher semantic level concept such as "chase" and "discuss" rather than low-level verbs like "run" or "speak". For event participants, the descriptions tend to be more detailed. For instance, a participant may be referred to as "a black-haired woman wearing a blue coat" rather than simply "a woman". 
%While providing richer information, it also makes event prediction more challenging.
% High-level verbs inherently convey more directional and complex concepts, often involving greater contextual nuances or situational intricacies, thereby necessitating deeper background knowledge or understanding for accurate prediction.
\textbf{Higher-order Markov chains} \cite{ky2018higher, zhao2021reviewvideopredictiveunderstanding}. Event prediction can be modeled as a higher-order Markov chain, in which the influence of historical events on future ones does not necessarily diminish with increasing temporal distance. In contrast, action prediction and video frame generation are often highly dependent on adjacent time steps. 
%For example, replacing \emph{EventB} in the event chain \emph{EventA-EventB-EventC-EventD-EventE} with \emph{EventF} could result in a change in the future event \emph{EventE} to \emph{EventG}, whereas this is less likely to occur in action prediction.
%Taking the scenario in Figure\ref{fig:top} as an example, replacing \emph{Event1: girl in pink talk with bald guy} with \emph{Event1: girl in pink argue with bald guy} could make the future event \emph{EventT: bald guy kisses girl in pink} less likely to happen.
Therefore, we believe that incorporating event prediction into video models is highly meaningful.

Compared with the existing event prediction tasks in NLP, video event prediction differs in following aspects: 
\textbf{Ambiguous coreference relationship} \cite{Liang2025videvent, wang-etal-2021-coreference}. In text events, objects are typically referred to by the same noun or pronoun, making them easily identifiable within the context. However, in video events, due to the changes in camera angles and variations in appearance, coreference relationships are often difficult to capture.
\textbf{More complex logic} \cite{wang2025fosteringvideoreasoningnextevent, helff2024vloldiagnosticdatasetvisual}. The involvement of different entities in various events makes event prediction tasks non-first-order logical, posing significant challenges to reasoning. For instance, in pairs of events such as ["Mr. Bean chases the thief" , "The thief escapes"] and ["Mr. Bean chases the thief" , "Mr. Bean fails to catch the thief"], the latter can be simplified to ["chase", "fail"], while the non-first-order logic means that the subject of the event may change in chains.

To fill in the gap in event prediction within the field of computer vision, in this paper, we introduce a novel video prediction task, which we term \emph{action-centric video event prediction (AVEP)} as shown in Figure \ref{fig:top}. We drew inspiration from the MCNC \cite{granroth2016happens} task in NLP, as well as action anticipation tasks \cite{grauman2022ego4d, damen2020epic} in computer vision, to design the AVEP task in a more scientifically grounded manner. To better address the semantic richness of video events, we propose to represent a video event in the form of a multimodal graph. 
% Thus, the task can be formulated as predicting the arguments of the next potential event, given a series of observed video events represented as graph structures.
Instead of following the MCNC task, we no longer provide the model with predefined options but directly predict the events. While this significantly increases the difficulty of the task, it aligns more closely with the original intent of event prediction. 

To support the proposed task, we have constructed the AVEP dataset, which contains around $35K$ videos with $178K$ video events, based on existing video event datasets including VidSitu\cite{sadhu2021visual}, Epic-kitchens\cite{damen2020epic} and VidEvent\cite{Liang2025videvent}. For each video, based on the textual descriptions of the events, we identify the event arguments and their corresponding image slices through a novel annotation procedure involving different labeling steps and a validation stage, and represent these in the form of event graphs. 

Considering existing visual models are based on frames or patches that are not well suited for the event structure, we propose an approach that adapts to the event graph structure by introducing a node-graph hierarchical attention Transformer to capture the relationships between historical events at the event graph level and the argument node level. This enables the model to analyze the factors that influence future events based on past occurrences, thus facilitating accurate predictions. Furthermore, to address the ambiguity of coreference relationships in video events, we introduce a novel coreference encoding method to represent the same object appearing across different video events. This enables the model to effectively recognize coreferential objects across temporal dimensions. Finally, we pre-train the model using random masking across the entire dataset, followed by post-training for the AVEP task. This training strategy enhances the model’s ability to capture and understand the relationships between events.

The contributions of this paper are as follows: 1) We introduce the action-centric video event prediction (AVEP) task, which fills in the gap in event prediction in computer vision. 2) We introduce the AVEP dataset, built upon several existing datasets, which contains approximately $35K$ videos along with $178K$ events annotated through a novel procedure to construct event graphs. 3) We propose a node-graph hierarchical attention and co-reference encoding mechanism to address the challenges in video event prediction. 4) Our proposed method outperforms existing SOTA models in related prediction tasks and open-source LVLMs in visual reasoning, establishing a strong baseline for the AVEP task.

\section{Related Work}
\label{sec:related_}

In this section, we will introduce other related work that has been mentioned previously in the following sections.
\subsection{Script Induction}
The concept of script knowledge in Artificial Intelligence, along with early knowledge-based methods to learn scripts were introduced by Schank and Abelson (1977) \cite{schank1975scripts} and Mooney and DeJong (1985) \cite{mooney1985learning}. 
% As shown in Fig1[], a script refers to a kind of structured knowledge, involving sequences of events. Scripts encode world knowledge that can help text understanding and script induction is to automatically learn this knowledge from unstructured text and predict the next occurring event.
One particularly influential work in this area is that of Chambers and Jurafsky (2008) \cite{chambers2008unsupervised}, which formulates the script induction problem as learning coherent event chains and predicting the missing event. Several follow-up works (Chambers and Jurafsky, 2009 \cite{chambers2009unsupervised}; Jans et al., 2012 \cite{jans2012skip}; Pichotta and Mooney, 2014 \cite{pichotta2014statistical}; Rudinger et al., 2015 \cite{rudinger2015learning}) employ and extend Chambers’s methods for learning narrative chains. Considering the uncertainty in the prediction results, Granroth-Wilding and Clark \cite{granroth2016happens} proposed a multiple-choice variation, called MCNC, to simplify the evaluation process. To address these challenges, numerous studies have proposed effective solutions. 
Seminal works \cite{yang2020nargnn, zheng2020heterogeneous} utilize graphs to structurally represent events instead of only texts, and some \cite{yang2020nargnn} leverage attention mechanisms to analyze the influencing factors of historical events. However, most of these studies only focus on the text modality and do not extend to the visual modality, while most events appear in form of videos in the real world. Hence, exploring event prediction in the visual domain is highly meaningful.

\subsection{Video Action Prediction}
In the realm of vision, several similar prediction tasks such as Action Anticipation, VLEP \cite{liu2020forecasting}, and VidEvent \cite{Liang2025videvent} have already been proposed.
These tasks are fundamental to many real world applications in area like robots and transportation \cite{kong2022human}. While these studies have made significant contributions to the field of video event prediction, there remains ample room for further research and improvement, particularly in terms of modeling event structures and designing more effective task formulations.

% Video action prediction task can be categorized by the perspective of the input video—either from a third-person view or from a first-person (egocentric) view.

Several excellent datasets have been developed to facilitate these video prediction tasks. VidSitu \cite{sadhu2021visual} is a large-scale dataset containing diverse videos from movies depicting complex situations (a collection of related events). VLEP \cite{lei2020more} is a multimodal commonsense next-event prediction dataset. 
VidEvent \cite{Liang2025videvent} is a dataset designed to advance the understanding of complex event structures in videos. 
Ego4D \cite{grauman2022ego4d} contains thousands of hours of egocentric video footage recorded in varied environments. EPIC-KITCHENS \cite{damen2020epic} is a large-scale dataset focusing on kitchen-based activities. However, although these datasets cover videos from various types of activities, they lack structured annotations describing the events within the videos, making them unsuitable for event prediction tasks. Therefore, a dataset specifically designed for detailed and structural descriptions of video events is needed.

In terms of recent model algorithms, many researchers have explored various models to address these video prediction tasks. 
% Vondrick\cite{vondrick2016anticipating} built a deep regression network that takes a single frame as input and anticipates the future frame representation of a pre-trained AlexNet. However, a single frame struggle to represent the temporal information of video events. 
Girdhar \cite{girdhar2021anticipative} presents a new model called Anticipative Video Transformer and a self-supervised future prediction loss for action anticipation. And Liang \cite{Liang2025videvent} proposes to use two separate Transformers to model video events and textual events respectively. 
% While action anticipation methods typically take the original video frames as the default input modality, some methods leverage higher level modalities like the camera wearer’s trajectory\cite{park2016egocentric}, hand trajectory\cite{liu2020forecasting}, eye gaze\cite{li2018eye}, and environment affordance\cite{nagarajan2020ego}. 
Recently, some \cite{kim2024palm} start to use large language models to help predict the actions in text modality. 
Existing prediction methods have demonstrated the effectiveness of the Transformer architecture in predicting action sequences. However, most current approaches are not well suited for the inherent structural nature of events. Therefore, we modify a node-graph hierarchical attention Transformer architecture to incorporate event structure into the model. 

\subsection{Graph-based Representations}
Compared to caption-based anticipation methods \cite{mun2017text, wang2023caption} that produce unstructured textual descriptions, graph-based representations offer explicit and compositional modeling of event structure and have been widely explored for video understanding \cite{qiu2025stepenhancingvideollmscompositional}. The Action Genome dataset \cite{ji2020action} decomposes each action into a spatio-temporal scene graph of objects and their relationships, and the Egocentric Action Scene Graphs (EASG) \cite{rodin2024action} extends this idea to first-person videos by providing temporally evolving graphs of verbs and interacting objects. Building on these representations, prior methods have applied GNNs and attention mechanisms for fine-grained event reasoning. ViGAT \cite{gkalelis2022vigat} employs Graph Attention Network blocks to model spatial and temporal dependencies among objects, while dynamic scene graph generation frameworks like TEMPURA \cite{cheng2025tempura} leverage transformer-based sequence modeling to learn object interactions over time and mitigate long-tail biases in video relation data. %And STEP \cite{qiu2025stepenhancingvideollmscompositional} utilizes a graph representation for video to enhance the reasoning capacity of LVLMs. 
However, these works focus on recognizing or detecting structured graphs in observed videos, whereas our approach aims to predict future event graphs by using the node-graph hierarchical attention to capture the relationships between historical events.

\section{AVEP Task}
\label{sec:AVEP Task}
In this section we formulate the AVEP task, which aims to predict the next potential event based on a series of observed historical video events. We extensively draw upon the definitions and evaluation frameworks of the natural language event task MCNC \cite{granroth2016happens} and video prediction-related tasks \cite{Liang2025videvent, liu2020forecasting} to ultimately define the form of the video event prediction task.

\subsection{Formal Definition}
We provide the following definition to describe the task:\\
\textbf{Definition 1} (Video Event). 
In the area of NLP, an event is typically defined as a occurrence that involves one or more participants, occurring in a specific time and place. 
%Events are often described by their arguments, which include subject, object, location and other contextual details. 
We adopt an NLP-inspired perspective to define a video event. A video event $E$ is characterized as an action or occurrence within a video clip $V$ containing $T$ frames, involving one or more participants. The video clip is treated as an atomic unit—further division would lead to the loss of essential event arguments. We retain the original definition of an event as Chambers \cite{chambers2008unsupervised}, in which an event $E$ can be represented as $E=(trg_t, ARG_t)$, where $trg_t$ is a word or span expressing the event and $ARG_t$ denotes the arguments of the event. To incorporate video-based descriptions of events, we extend $trg_v$ and $ARG_v$ using frames or frame slices related with the arguments, $f_{trg}$ and $f_{ARG}$, to describe the event's trigger and the corresponding arguments, in which $trg_v=(f_{trg0},f_{trg1},\dots, f_{trgn};trg_t)$ and $ARG_v=(f_{ARG0},f_{ARG1},\dots, f_{ARGn};ARG_t)$.\\
\textbf{Definition 2} (Event Graph). 
To better represent the structural relationships between event arguments, we introduce a directed graph structure to represent an event, called an event graph $\mathcal{G}_E=(\mathcal{V},\mathcal{E})$, as shown in Figure \ref{fig:top}. In such a graph, the nodes $\mathcal{V}=\{trg_v;ARG_v\}$ represent the event trigger and arguments, and the edges $e=rel(v_i,v_j)$ represent the relationships between these arguments. Here, $v\in \mathcal{V},e\in \mathcal{E}$, with the function $rel$ is designed to define the argument roles of nodes in an event, such as $subj.$, $obj.$, and $loc.$. \\
\textbf{Definition 3} (Event Chain). 
Followed Granroth-Wilding and Clark \cite{granroth2016happens}, in which the observed historical event sequence is represented in the form of an event chain $C=(E_1,E_2,\dots,E_n)$, we define an event chain through a sequence of event graphs $\mathcal{C}=(\mathcal{G}_{E_1},\mathcal{G}_{E_2},\dots,\mathcal{G}_{E_n})$. 

\subsection{Action-centric Video Event Prediction Task Definition}
As shown in Figure \ref{fig:top}, its formalized expression is as follows: Given an event graph chain $\mathcal{C}=(\mathcal{G}_{E_1},\mathcal{G}_{E_2},\dots,\mathcal{G}_{E_n})$ representing a sequence of observed historical events, AVEP requires the model to predict $trg_t$ and $ARG_t$ within the possible future event. Each event graph chain $\mathcal{C}$ consists of $t$ event graphs $\mathcal{G}_{E_i}=(\mathcal{V}_i,\mathcal{E}_i)$, where each event graph is composed of $m$ multimodal argument nodes $v_{ij}\in \mathcal{V}_i$ and their corresponding edges $e_{ij}\in \mathcal{E}_i$. To more accurately evaluate the model's performance in real-world applications, we no longer provide five candidate future events as in the MCNC task. Instead, following the design of the action anticipation task, we require the model to directly predict the future event's $trg_t$ from the verb vocabulary of the entire dataset and further predict the arguments $ARG_t$ associated with $trg_t$. 

\section{Dataset}
\label{sec:dataset}

We expand existing event-level video datasets to construct AVEP dataset, a large-scale dataset for video event prediction, which structurally represents video events using event graphs. 
Moreover, we divide the AVEP into first-person perspective and third-person perspective to cater to the needs of different downstream tasks and real-world applications.
%%%%
%First-person and third-person perspectives are commonly defined as the videos recorded from the viewpoint of the subject themselves and the third-person perspective refers to videos captured from an external perspective \cite{sigurdsson2018charades}. 

The proposed dataset meets the requirements of video event prediction tasks in terms of task support, semantic richness, structural clarity, and practical applicability. We further describe the detailed data and data analysis, along with the overview of the annotation process which involves different labeling steps and a validation stage.

% We also provide a detailed description of all the datasets utilized in the expansion process to construct the AVEP dataset in the Appendix \ref{sec:apA}.
We also provide a detailed description of all the datasets utilized in the expansion process to construct the AVEP dataset in the Appendix.

\subsection{Construction Procedure}
Our dataset construction process consists of three main stages: data collection, integration and expansion, and manual verification. First, we conducted a comprehensive survey of existing event-related video datasets, evaluating them based on their structural support for event representation and their suitability for event prediction tasks. Based on this evaluation, we selected appropriate datasets as the foundation for constructing the AVEP dataset.

In the integration and expansion stage, we standardized all textual annotations across datasets into a unified event-structured format. Additionally, we developed an automatic argument annotation tool based on the GDINO \cite{liu2024grounding} model to generate bounding box annotations for event arguments in videos, followed by cropping the corresponding regions.

% Finally, during the data verification stage, three annotators reviewed and refined the automatically generated argument annotations to ensure accuracy and consistency. A detailed description of the dataset construction process is provided in the Appendix \ref{sec:apA}.
Finally, during the data verification stage, three annotators reviewed and refined the automatically generated argument annotations to ensure accuracy and consistency. A detailed description of the dataset construction process is provided in the Appendix.

\subsection{Data Statistics}
% AVEP dataset contains more than $178K$ video event graphs of complex structures, rich hierarchy, and logical evolutionary. As shown in Table \ref{tab:count} in Appendix \ref{sec:apA}, we provide statistics on the splits of the dataset. To validate these characteristics, we performed a detailed statistical analysis of the dataset from multiple perspectives.

AVEP dataset contains more than $178K$ video event graphs of complex structures, rich hierarchy, and logical evolutionary. We provide statistics on the splits of the dataset in Appendix. To validate these characteristics, we performed a detailed statistical analysis of the dataset from multiple perspectives.

In our dataset, event chains range in length from 3 to 15, which is sufficient to cover most event reasoning scenarios. Furthermore, each event involves 2 to 5 arguments, with an average of $2.8$ arguments per event, aligning well with the definitions used in previous studies.

The dataset comprises a total of $2284$ unique verbs. Our analysis of the verb distribution reveals a relatively balanced spread. Notably, the top 30 most frequently occurring verbs collectively constitute less than 40\% of the entire set, with no single verb overwhelmingly dominating the distribution. This level of lexical diversity ensures comprehensive coverage of the vast majority of events encountered in real-world applications.

To evaluate whether the dataset sufficiently captures the rich semantics of events, we analyzed the distribution of nouns in the annotations of $ARGs$. The results indicate that the dataset contains over 6,000 unique nouns. Excluding commonly used terms such as "man" and "woman", which frequently appear in annotations referring to people, we found that the remaining nouns are relatively evenly distributed. This balanced distribution ensures that our dataset aligns with the semantic richness observed in real-world applications. 

% We provide a detailed dataset statistics in the Appendix \ref{sec:apA}.
We provide a detailed dataset statistics in the Appendix.

\section{Method}
\label{sec:method}

In this section, we will provide a detailed introduction of the EventFormer model designed for video event prediction, along with the baseline models we developed for the AVEP task based on this framework. 

\subsection{EventFormer}
In a standard Transformer model, the basic unit of input is the tokens, and all tokens operate at the same hierarchical level within the architecture. 
For the AVEP task, the model needs to attend not only to the key events in the historical event chain, but also to the key arguments within historical events. 
However, taking either node or graph embeddings as tokens, this single-level structure fails to capture the complex relationships between nodes and graphs at the same time. 
To address this limitation, we propose EventFormer, a temporal prediction framework incorporating a node-graph hierarchical attention mechanism tailored for graph-structured data, as shown in Figure \ref{fig:attention}.  

\subsubsection{Overview}
We basically follow the architecture of Transformer \cite{vaswani2017attention} to construct EventFormer, while extending a novel \textbf{Node-graph Hierarchical Attention} mechanism to adapt to the complex structure of events.
The EventFormer is composed of a stack of $N$ layers, in which each layer has two sub-layers, taking an event graph chain $\mathcal{C}$ as input. The first is a Node-graph Hierarchical Attention mechanism module network, and the second is a position-wise fully connected feed-forward network. Following the Transformer encoder architecture, we similarly employ a residual connection \cite{he2016deep} around each of the two sub-layers, followed by layer normalization. To facilitate these residual connections, all sub-layers in the model, as well as the embedding layers, produce outputs of dimension $d$. 

\subsubsection{Node-graph Hierarchical Attention.}
\begin{figure}
    \centering
    \includegraphics[width=1\linewidth]{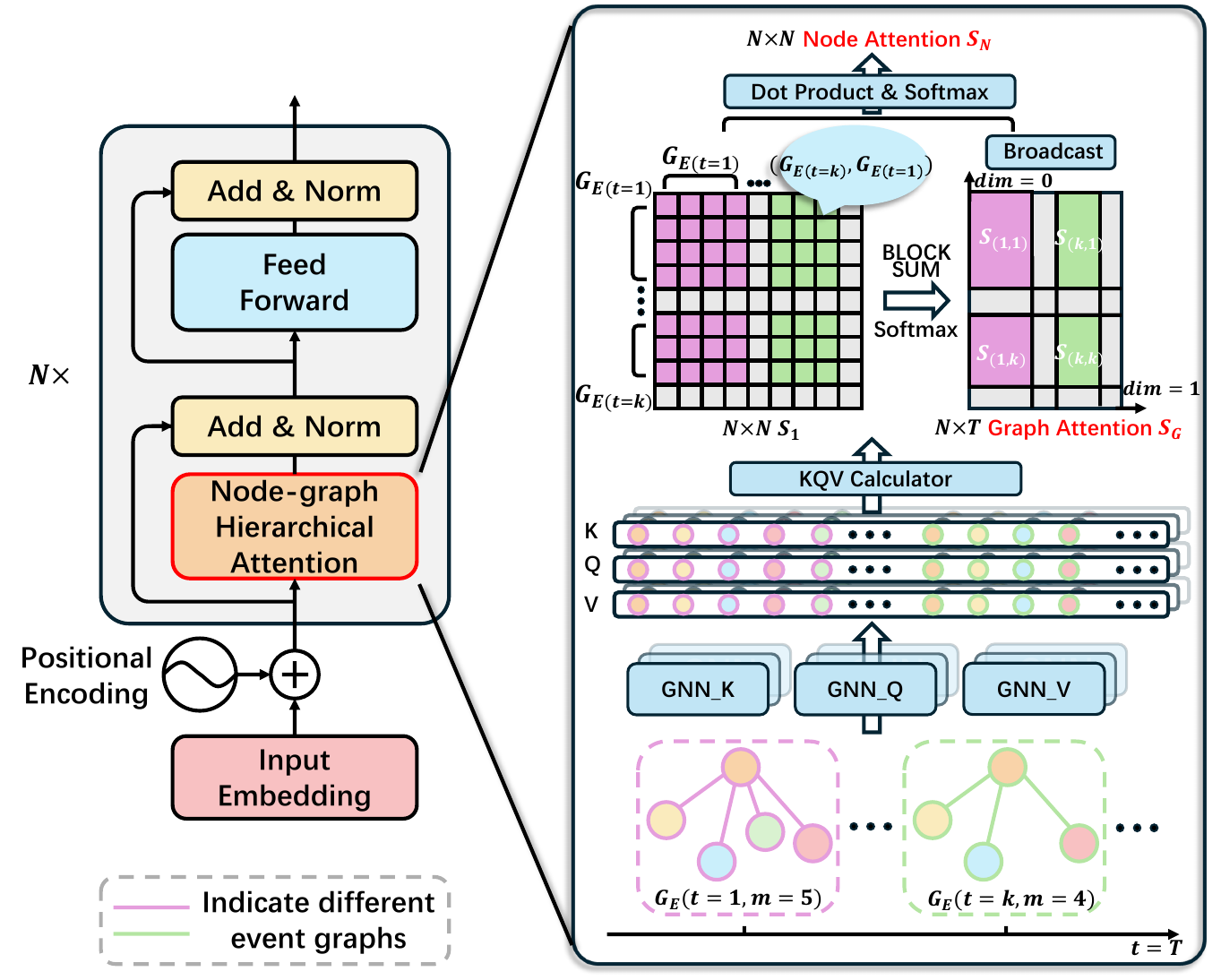}
    \caption{The architecture of EventFormer. On the left is the overall model framework, which follows the same architecture as Transformer. On the right is a magnified view of Node-graph Hierarchical  Attention, providing detailed explanations of its implementation. We include calculation examples for these components in the Appendix.}
    \label{fig:attention}
\end{figure}

\begin{figure*}[ht]
    \centering
    \includegraphics[width=1.0\linewidth]{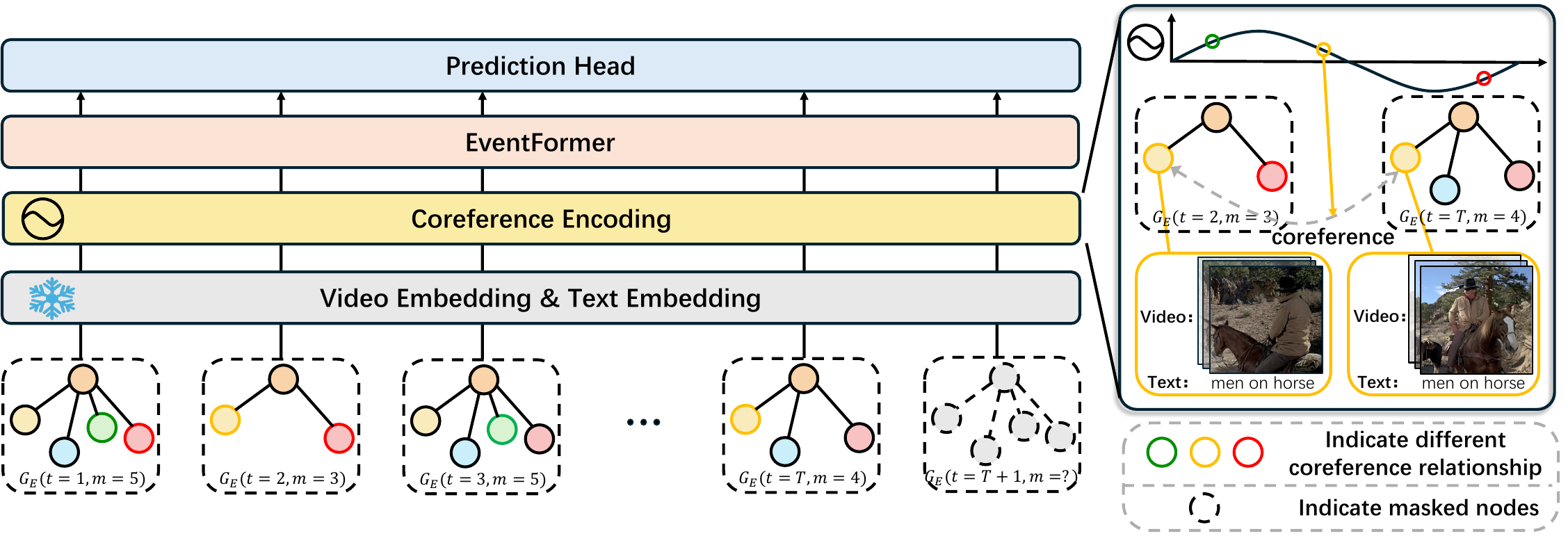}
    \caption{The architecture of Video Event Prediction Model based on EventFormer. The left side of the image shows the overall framework of the model. The frozen symbols indicate that the model's parameters are frozen during training. The right side of the image provides an detailed view of the Coreference Encoding module. The same colored borders represent coreference relationships, while the dashed borders indicate that future events are replaced by the learned mask.}
    \label{fig:model}
\end{figure*}

To better accommodate the structure of events, we introduce the node-graph hierarchical attention mechanism, as shown in Figure \ref{fig:attention}. This mechanism enables the model to capture the attention at both the node level and the graph level, including node-to-node interactions within the same graph, node-to-node interactions across different graphs, and the interactions between different graphs. 

We treat an event graph $G_E$ as a token, which is composed of $m$ units of arguments, while applying several \textbf{graph neural networks}(GNN) to generate the $KQV$ values instead of linear layers. This allows the model to build a more comprehensive event representation by capturing both internal argument interactions and the relationships between them.
\begin{equation}
    K=GNN_K(G_E),Q=GNN_Q(G_E),V=GNN_V(G_E), KQV\in R^{N\times d},
\end{equation}
where $N$ represents the number of all nodes in $\mathcal{C}$ and $d$ denotes the feature dimension. The GNN can be substituted with various basic graph neural network models \cite{kipf2016semi, velickovic2018graph, xu2018powerful}, and we will compare the performance of these models in the Experiment Section.

With the $KQV$ generated by multi-head GNNs, we apply the same attention score method at the node level, as follows:
\begin{equation}
    S_1=\text{softmax}(\frac{QK^T}{\sqrt{d}}),S_1\in R^{N\times N}.
\end{equation}
We further partition the $S_1$ according to the sequence of graphs in $\mathcal{C}$, and assemble them into blocks to represent the node level attention scores between each graph pairs $(G_{E(t=i)},G_{E(t=j)})$, which indicates the node attention scores of $G_{E(t=j)}$ towards $G_{E(t=i)}$. Then we apply a $\text{BlockSum}$ function on the blocks to generate a $N\times T$ $S_G$, representing the Graph Attention between graph pairs.
\begin{equation}
    S_G=\text{softmax(}\text{BlockSum}(S_1);dim=1),S_G\in R^{N\times T}.
\end{equation}
After that, we use $S_G$ to compute Node Attention $S_N$ as follows:
\begin{equation}
    S_N=\text{softmax}(S_1\cdot \text{Broadcast}(S_G)),S_N\in R^{N\times N},
\end{equation}
in which we employ a broadcasting mechanism to expand $S_G$ to the same dimension as $S_1$ based on the number of nodes in each graph, enabling element-wise multiplication.
With the attention score $S_N$ and the calculated $V$, we update the node features $\mathcal{F}$ of each event graph $G_{E(t)}$ in chain $\mathcal{C}$ with the latent representation generated by the Node-graph Hierarchical Attention module,
\begin{equation}
    \mathcal{F}=S_N\times V,\mathcal{F}\to\mathcal{C}.
\end{equation}
 
\subsection{Video Event Prediction Model}
In this section, we elaborate on how EventFormer is applied to the video event prediction task, where the model predicts future event graphs $\hat{G_E}$ based on the input event graph chain $\mathcal{C}$. Additionally, we introduce a novel coreference encoding mechanism to address the ambiguity in coreference relationships within video events.

\subsubsection{Overview}

As shown in Figure \ref{fig:model}, the Video Event Prediction Model is composed of four main components: Video and Text Embedding, Coreference Encoding, EventFormer and Prediction Head. We utilize a pre-trained CLIP \cite{radford2021learning} model as an embedding model and keep its parameters frozen during training. For each event graph $G_{Ei}$, we utilize the CLIP model to generate textual features $F_T$ and video features $F_V$ for each node $v_{ij}=(f_0,f_1,\dots,f_n;trg/ARG)$, in which we apply pooling to the extracted frame representations to generate the video features. We represent the features of nodes $F_{ij}$ in event graph by concatenating the textual and video features:
\begin{equation}
    F_{ij}=\text{Concat}(F_{Tij},F_{Vij}),
\end{equation}
in which $i$ denotes the graph number and $j$ denotes the node number. Then we introduce a coreference encoding mechanism to capture coreferential relationships within $\mathcal{C}$, enhancing the model's ability to distinguish coreferential nodes. This mechanism $ \text{CE}$ will be explained in detail in the next subsection:
\begin{equation}
F_{ij}' = \text{CE}(F_{ij}) + F_{ij}.
\end{equation}
At this point, we obtain the multimodal features for each node in the video event chain. The $\mathcal{C}$ are then input into the EventFormer model, yielding a hidden representation $\mathcal{C\prime}$ of the event chain in the feature space. Using this hidden representation, we predict future event $\hat{G_{E}}$ through a prediction head composed of multiple fully connected layers. 
As shown in Figure \ref{fig:model}, to enable event prediction, we apply a learnable graph mask to conceal future events within the event chain as $G_{masked}$. For prediction, the prediction head of model $\text{Head}$ utilizes the node features of the masked graph to generate the logits of verb $logits_{verb}$ and the predicted noun embeddings $E_{noun}$,
\begin{equation}
    logits_{verb},E_{noun}=\text{Head}(G_{masked}).
\end{equation}

\subsubsection{Coreference Encoding}
Event graph nodes with coreferential relationships typically share identical textual annotations, yet they may exhibit significant differences in the visual modality. For instance, as shown in Figure \ref{fig:model}, the same "man on a horse" appears in two different event graphs. However, due to variations in scene and perspective, it is not immediately evident from the video alone that they refer to the same individual.

Therefore, we introduce a novel coreference encoding mechanism to equip the model with the ability to capture the coreference relationships between nodes. Before feeding the video event chain $\mathcal{C}$ into the EventFormer model, we incorporate coreference encodings into embeddings of nodes with coreferential relationships. The corefernece encodings have the same dimension $d$ as the node embeddings, so that the two can be summed. There are many choices of coreference encodings, learned and fixed \cite{su2023roformerenhancedtransformerrotary, gehring2017convolutionalsequencesequencelearning, vaswani2017attention}. In this work, we use sine and cosine functions of different frequencies:
\begin{align}
    \text{CE}_{(index, 2k)}=s\cdot sin(index/10000^{2k/d}),\\
   \text{CE}_{(index, 2k+1)}=s\cdot cos(index/10000^{2k/d}),
\end{align}
in which $index$ denotes the index of $ARGs$ in the video event chain $\mathcal{C}$, $k$ denotes the embedding dimension and $s$ denotes the scaling factor of the coreference encoding. By directly adding the coreference encodings to node embeddings, we obtain a video event chain that incorporates coreferential relationships. We demonstrate the effectiveness of the coreference encoding through experiments, as shown in the Table \ref{tab:ce} in the Experimental Section.

\subsection{Training}
This section describes the training regime for our models.

\textbf{Loss Function.} 
To predict verbs and nouns in future video events, we employ a combination of weighted cross-entropy loss, focal loss, and mean squared error (MSE) loss, denoted as $\mathcal{L}_{ce}$, $\mathcal{L}_{focal}$, and $\mathcal{L}_{mse}$, respectively. For verb prediction, we utilize cross-entropy loss, formulated as:
\begin{equation} 
    \mathcal{L}_{ce}=\text{CrossEntropy}(logits_{verb}, gt), 
\end{equation}
where $gt$ represents the ground truth labels. Given that the target nouns to be predicted have appeared in past events, we leverage the similarity score $s$ between the predicted noun embeddings and historical noun embeddings to identify noun nodes in future events. This similarity-based approach is incorporated into the focal loss to enhance noun prediction:
\begin{gather} 
    s=\text{CosineSimilarity}(E_{noun}, ARGs),\\ \mathcal{L}_{focal}=\text{FocalLoss}(s, gt). 
\end{gather}
Furthermore, to ensure that the predicted noun embeddings closely approximate the ground truth noun embeddings, we introduce an additional MSE loss term.
Finally, the overall objective function is defined as a weighted sum of the three loss components:
\begin{equation} 
    \mathcal{L}=\alpha\mathcal{L}_{ce}+\beta\mathcal{L}_{focal}+\gamma\mathcal{L}_{mse}. 
\end{equation}

\begin{table*}[t]
\centering
\caption{The comparative experiment with other SOTA models. }
\label{tab:comparative}
\setlength{\tabcolsep}{2.9mm}{
\begin{tabular*}{\linewidth}{l|l|l|l|ll|ccc|ccc}
\toprule[1pt]
\multirow{2}{*}{Perspective}  & \multirow{2}{*}{Set}  & \multirow{2}{*}{Method} & \multirow{2}{*}{Parameter} & \multicolumn{2}{c|}{Verb} & \multicolumn{3}{c|}{Noun} & \multicolumn{3}{c}{Verb-Noun}\\ \cline{5-12} 
    &     &   &  & Top1 & Top5 & P & R & F1 & P & R & F1 \\ 
\midrule
\multirow{8}{*}{third person} & \multirow{4}{*}{val}  
& VidEvent & 44.1M & 18.97 & \textbf{45.61} & 32.62 & 34.39 & 33.48 & 6.19 & 6.45 & 6.32\\
&   & LLaVA-Video & 7B & 5.67 & 21.32 & 44.50 & 41.95 & 43.19 & 3.13 & 2.95 & 3.04\\ 
&   & Qwen2.5-VL & 72B & 2.05 & 16.82 & \textbf{52.74} & \textbf{51.71} & \textbf{52.22} & 1.43 & 1.40 & 1.41\\
&   & Ours & \textbf{7.3M} & \textbf{22.32} & 45.16 & 33.10 & 44.90 & 38.11 & \textbf{7.67} & \textbf{10.88} & \textbf{9.00}\\ \cline{3-12}
&   & Human* & - & 37.42 & 60.65 & 61.05 & 55.18 & 57.97 & 25.09 & 21.04 & 22.89\\
\cline{2-12} & 
\multirow{4}{*}{test} 
& VidEvent & 44.1M & 18.44 & 43.96 & 36.49 & 37.06 & 36.77 & 5.85 & 5.91 & 5.88\\
&   & LLaVA-Video & 7B & 6.27 & 10.30 & 44.96 & 41.64 & 43.24 & 3.57 & 3.30 & 3.43  \\ 
&   & Qwen2.5-VL & 72B & 2.00 & 16.23 & \textbf{52.23} & \textbf{54.31} & \textbf{53.25} & 1.21 & 1.26 & 1.23\\
&   & Ours & \textbf{7.3M} & \textbf{22.71} & \textbf{45.21} & 42.93 & 50.11 & 46.24 & \textbf{7.67} & \textbf{7.71} & \textbf{7.69} \\ \cline{3-12}
&   & Human* & - & 38.46 & 63.08 & 64.80 & 52.22 & 57.83 & 27.15 & 22.22 & 24.44\\
\midrule
\multirow{3}{*}{first person} & \multirow{4}{*}{test} 
& InAViT & 157.2M & 29.78 & 64.00 & - & 26.67 & - & - & \textbf{8.69} & -  \\
&   & LLaVA-Video & 7B & 3.49 & 5.23 & \textbf{25.42} & 45.50 & 32.62 & 1.40 & 2.50 & 1.79\\   
&   & Qwen2.5-VL & 72B & 1.16 & 25.34 & 23.56 & \textbf{55.66} & \textbf{33.11} & 0.46 & 1.08 & 0.65 \\
&   & Ours & \textbf{7.3M} & \textbf{29.94} & \textbf{65.25} & 18.88 & 23.35 & 20.88 & \textbf{6.43} & 7.67 & \textbf{7.00}\\
\bottomrule[1pt]
\end{tabular*}}
\end{table*}

\textbf{Two-stage Training.} 
We adopt a two-stage training strategy to train our Eventformer on AVEP task. We first apply a random masking approach to the input event graph sequence, randomly masking certain event graphs and training the model to reconstruct them during the pre-training phase. This pre-training strategy enables the model to better capture relationships between events and enhances its reasoning ability for event inference. Then we proceed with a post-training phase, where the pre-trained model undergoes a few additional training epochs on the specific task. In this phase, the masking is fixed on the last event graph in the sequence, allowing the model to be fine-tuned on AVEP. Table \ref{tab:pretrain} in Experimental demonstrates that this training strategy effectively improves the model’s performance on AVEP task.

\section{Experiments}

In this section, we employ SOTA models \cite{roy2024interactionregionvisualtransformer, Liang2025videvent} from existing prediction tasks, along with LVLMs\cite{lin2024videollavalearningunitedvisual, bai2025qwen25vltechnicalreport}, to conduct a series of comparative experiments, validating the complexity of our proposed task and the significance of our dataset. All LVLMs used in our experiments were evaluated in a zero-shot setting. In addition, we design both comparative and ablation studies to assess the effectiveness of our proposed approach. 

\subsection{Setups}
We use CLIP as our node feature embedding model, which is frozen when training. To assess the impact of different GNN architectures on performance, we performed experiments on GCN, GAT, and GIN implementations from the Deep Graph Library (DGL) \cite{wang2019dgl}. The dimension of hidden feature $d$ in our model is $1024$. 
Based on the experimental results, we set the loss weights $[\alpha,\beta,\gamma ]$ to $[1.0,1.0,0.5]$. Our model is optimized by Adam, and the learning rate and weight decay are $1e-5$ and $1e-6$. Setting the maximum training step to $300$, we select the performance of the best epoch as our final results. Dropout ratio and batch size are 0.3 and 64. The maximum sequence length is 50. Training is conducted on 2×3090 GPUs and takes $\sim3$ days. 

We measure future event verb prediction using the Acc@K. For noun argument prediction in future events, we evaluate performance using recall, precision, and F1-score metrics.
We also introduce a verb-noun metrics to to assess the model’s predictive capability in future event prediction, where the verb-noun recall, precision and F1-score represent the recall, precision and F1-score of noun prediction, conditioned on the correct prediction of the verb. 

\begin{table*}[t]
\centering
\caption{The ablation experiment on different GNNs. }
\label{tab:gnn}
\setlength{\tabcolsep}{3.45mm}{
\begin{tabularx}{1.0\linewidth}{c|c|l|cc|ccc|ccc}
\toprule[1pt]
\multirow{2}{*}{Perspective}  & \multirow{2}{*}{Set}  & \multirow{2}{*}{Method} & \multicolumn{2}{c|}{Verb} & \multicolumn{3}{c|}{Noun} & \multicolumn{3}{c}{Verb-Noun}\\ \cline{4-11} 
    &     &     & Top1 & Top5 & P & R & F1 & P & R & F1\\ 
\midrule
\multirow{8}{*}{third person} & \multirow{4}{*}{val} 
& EventFormer-Linear & 11.38 & 30.21 & 30.15 & 41.1 & 34.78 & 4.02 & 6.34 & 4.92\\
&   & EventFormer-GCN & 14.96 & 38.99 & 29.41 & 41.72 & 34.50 & 4.56 & 6.89 & 5.49\\
&   & EventFormer-GAT & 16.00 & 39.58 & 31.09 & \textbf{47.74} & \textbf{37.66} & 4.13 & \textbf{10.13} & 5.87\\
&   & EventFormer-GIN & \textbf{19.20} & \textbf{41.67} & \textbf{32.95} & 42.13 & 36.98 & \textbf{6.71} & 9.93 & \textbf{8.01}\\
\cline{2-11} & 
\multirow{4}{*}{test} 
& EventFormer-Linear & 15.31 & 36.35 & 38.53 & 39.23 & 38.88 & 3.92 & 4.22 & 4.06\\
&   & EventFormer-GCN & 14.97 & 39.79 & 36.34 & \textbf{49.86} & 42.04 & 4.47 & 5.02 & 4.73\\
&   & EventFormer-GAT & 13.61 & 40.76 & \textbf{42.55} & 47.51 & \textbf{44.89} & 4.13 & 4.67 & 4.38\\
&   & EventFormer-GIN & \textbf{19.20} & \textbf{42.33} & 41.68 & 48.62 & 44.88 & \textbf{6.58} & \textbf{9.67} & \textbf{7.83}\\
\bottomrule[1pt]
\end{tabularx}}
\end{table*}

\begin{table*}[]
\caption{The ablation experiment on Coreference Encoding. }
\label{tab:ce}
\setlength{\tabcolsep}{3.75mm}{
\begin{tabular*}{1.0\linewidth}{c|c|l|cc|ccc|ccc}
\toprule[1pt]
\multirow{2}{*}{Perspective}  & \multirow{2}{*}{Set}  & \multirow{2}{*}{Method} & \multicolumn{2}{c|}{Verb} & \multicolumn{3}{c|}{Noun} & \multicolumn{3}{c}{Verb-Noun}\\ \cline{4-11} 
    &     &     & Top1 & Top5 & P & R & F1 & P & R & F1\\ 
\midrule
\multirow{6}{*}{third person} & \multirow{3}{*}{val}  
& Ours-w/ 0.5CE & 15.92 & 40.03 & 32.30 & 38.99 & 35.33 & 4.62 & 8.88 & 6.08\\ 
&   & Oursr-w/o CE & 16.52 & 41.07 & 31.83 & 41.17 & 35.902 & 5.73 & \textbf{10.18} & 7.33\\
&   & Ours-w/ CE& \textbf{19.20} & \textbf{41.67} & \textbf{32.95} & \textbf{42.13} & \textbf{36.98} & \textbf{6.71} & 9.93 & \textbf{8.01}\\
\cline{2-11} & 
\multirow{3}{*}{test} 
& Ours-w/ 0.5CE & 16.02 & 41.05 & 34.71 & 39.44 & 36.92 & 4.65 & 8.63 & 6.04\\ 
&   & Ours-w/o CE & 16.96 & \textbf{43.23} & 40.22 & 42.53 & 41.34 & 6.33 & 7.62 & 6.92\\
&   & Ours-w/ CE & \textbf{19.20} & 42.33 & \textbf{41.68} & \textbf{48.62} & \textbf{44.88} & \textbf{6.58} & \textbf{9.67} & \textbf{7.83}\\
\bottomrule[1pt]
\end{tabular*}}
\end{table*}

\begin{table*}[]
\caption{The ablation experiment on two-stage training. }
\label{tab:pretrain}
\setlength{\tabcolsep}{3.35mm}{
\begin{tabular*}{1.0\linewidth}{c|c|l|cc|ccc|ccc}
\toprule[1pt]
\multirow{2}{*}{Perspective}  & \multirow{2}{*}{Set}  & \multirow{2}{*}{Method} & \multicolumn{2}{c|}{Verb} & \multicolumn{3}{c|}{Noun} & \multicolumn{3}{c}{Verb-Noun}\\ \cline{4-11} 
    &     &     & Top1 & Top5 & R & P & F1 & R & P & F1\\ 
\midrule
\multirow{6}{*}{third person} & \multirow{3}{*}{val}  
& Ours(one-stage train) & 19.20 & 41.67 & 32.95 & 42.13 & 36.98 & 6.71 & 9.93 & 8.01\\
&   & Ours(two-stage train) & 22.32 & 45.16 & 33.10 & 44.90 & 38.11 & 7.67 & 10.88 & 9.00\\ 
\cline{4-11}
&   & Improvement & +3.12 & +3.49 & +0.15 & +2.77 & +1.13 & +0.96 & +0.95 & +0.99\\
\cline{2-11} & 
\multirow{3}{*}{test} 
& Ours(one-stage train) & 19.20 & 42.33 & 41.68 & 48.62 & 44.88 & 6.58 & 9.67 & 7.83\\
&   & Ours(two-stage train) & 22.71 & 45.21 & 42.93 & 50.11 & 46.24 & 7.67 & 7.71 & 7.69\\
\cline{4-11}
&   & Improvement & +3.51 & +2.88 & +1.25 & +1.49 & +1.36 & +1.09 & -1.96 & -0.14 \\
\bottomrule[1pt]
\end{tabular*}}
\end{table*}
\vspace{-3pt}
\subsection{Comparative Experiments}
% Considering the inherent ambiguity in AVEP, we provide human evaluation results as a credible reference point for assessing model performance. To this end, we simulate human performance on the task, offering a benchmark that reflects the upper bound of predictive accuracy. Further details are in the Appendix \ref{sec:apD}.
Considering the inherent ambiguity in AVEP, we provide human evaluation results as a credible reference point for assessing model performance. To this end, we simulate human performance on the task, offering a benchmark that reflects the upper bound of predictive accuracy. Further details are in the Appendix.

% Comprehensive details of the baseline models used for comparison are provided in the Appendix \ref{sec:apE}.
Comprehensive details of the baseline models used for comparison are provided in the Appendix.
As shown in Table \ref{tab:comparative}, for verb prediction, our proposed model significantly outperforms all other models in terms of ACC@1 by \textbf{4.27}, and achieves comparable performance to the best-performing model on ACC@5. Notably, the LVLM demonstrates poor performance in verb prediction, suggesting that current LVLMs still lack reliable capabilities for video-based event reasoning. 
For the prediction of noun arguments, our method outperforms \textit{VidEvent} by achieving improvements of \textbf{13.05} in Recall, \textbf{6.44} in Precision and \textbf{9.47} in F1-score. Interestingly, LVLMs exhibit strong performance in noun prediction. We attribute this to their robust visual recognition capabilities, which allow them to accurately identify and associate characters and objects appearing in the video. 
Compared to other metrics, the Verb-Noun metric provides a more comprehensive reflection of a model’s capability in future event prediction. Our proposed method achieves the best results on all three metrics across both the validation and test sets. 

We also evaluated our model’s performance on first-person AVEP. While our method still outperforms existing SOTA models in verb prediction, it does not surpass them in noun or verb-noun prediction. This is primarily because, to ensure a fair and realistic comparison, we strictly followed the input format of the original InAViT model, which includes the initial frames of the future events. As a result, the input frames may already contain visual clues about the upcoming nouns, making the noun prediction task easier.

\subsection{Ablation Experiments}
We conducted a series of ablation studies to systematically evaluate the effectiveness of the proposed method. 
\subsubsection{Ablation on GNN}
We conducted experiments on the third-person AVEP using different GNN architectures and linear layers. As shown in Table \ref{tab:gnn}, the linear layer performed the worst across all metrics, indicating that incorporating GNNs can effectively enhance token representation in graph-structured inputs. Among the three basic GNN models, GIN achieved the best overall performance on most metrics. In particular, it significantly outperformed GAT and GCN in verb prediction and verb-noun prediction from the event structure. Meanwhile, GAT showed competitive performance in noun prediction, suggesting its relative strength in capturing localized node-level information.
\subsubsection{Ablation on Coreference Encoding}
We conducted an ablation study on coreference encoding (CE) on the third-person AVEP, as shown in Table \ref{tab:ce}. Specifically, we evaluated three configurations: (1) without CE (2) with CE using a scaling factor of $0.5$, and (3) with a scaling factor of $1.0$. Among them, the model with CE at $scale = 0.5$ performed the worst. In contrast, introducing coreference encoding ($scale = 1.0$) led to consistent improvements across all evaluation metrics compared to the no-CE setting. Notably, verb prediction accuracy (ACC@1) improved by 2.68, and the F1-score for verb-noun prediction increased by 0.68. These results indicate that the proposed CE method helps the model better capture the semantic relationships among event arguments, thereby enhancing its ability to predict future events.
\vspace{-3pt}
\subsubsection{Ablation on Two-stage Training}
Finally, we conducted an ablation study on the two-stage training strategy, as shown in Table \ref{tab:pretrain}. Compared to training the model directly on the AVEP task, pretraining the model with a random masking strategy followed by task-specific post-training led to substantial performance improvements across most evaluation metrics. Although results on the test set show a slight drop in the verb-noun Precision and F1-score under the two-stage training setup, the significant gains in the majority of other metrics still demonstrate that this training strategy effectively enhances the model’s capability in future event prediction. 
\vspace{-10pt}
\section{Conclusion}
We introduce a novel task, Action-centric Video Event Prediction (AVEP), which requires models to predict the most probable future event trigger and its associated arguments based on a given video event chain. To facilitate research on this task, we construct the AVEP dataset based on several existing video datasets.
To address the challenges of this task, we propose a node-graph hierarchical attention Transformer coupled with a coreference encoding mechanism, which jointly captures the structural relationships among events and arguments as well as the coreferential dependencies between arguments across different events.
% Furthermore, we demonstrate through experiments that the two-stage training strategy is effective in improving event prediction performance.
Our method achieves SOTA performance on AVEP task, providing a strong baseline for further research.

\begin{acks}
This study is partially supported by National Natural Science Foundation of China (62176016, 72274127), 
National Key R\&D Program of China (No. 2021YFB2104800), 
Guizhou Province Science and Technology Project: Research on Q\&A Interactive Virtual Digital People for Intelligent Medical Treatment in Information Innovation Environment (supported by Qiankehe[2024] General 058), 
Capital Health Development Research Project(2022-2-2013), 
Haidian innovation and translation program from Peking University Third Hospital (HDCXZHKC2023203), 
and Project: Research on the Decision Support System for Urban and Park Carbon Emissions Empowered by Digital Technology - A Special Study on the Monitoring and Identification of Heavy Truck Beidou Carbon Emission Reductions.
\end{acks}
\bibliographystyle{ACM-Reference-Format}

\begin{thebibliography}{58}

%%% the Web address.

\ifx \showCODEN    \undefined \def \showCODEN     #1{\unskip}     \fi
\ifx \showISBNx    \undefined \def \showISBNx     #1{\unskip}     \fi
\ifx \showISBNxiii \undefined \def \showISBNxiii  #1{\unskip}     \fi
\ifx \showISSN     \undefined \def \showISSN      #1{\unskip}     \fi
\ifx \showLCCN     \undefined \def \showLCCN      #1{\unskip}     \fi
\ifx \shownote     \undefined \def \shownote      #1{#1}          \fi
\ifx \showarticletitle \undefined \def \showarticletitle #1{#1}   \fi
\ifx \showURL      \undefined \def \showURL       {\relax}        \fi
% invisible to TeX
\providecommand\bibfield[2]{#2}
\providecommand\bibinfo[2]{#2}
\providecommand\natexlab[1]{#1}
\providecommand\showeprint[2][]{arXiv:#2}

\bibitem[Abu-El-Haija et~al\mbox{.}(2016)]%
        {abu2016youtube}
\bibfield{author}{\bibinfo{person}{Sami Abu-El-Haija}, \bibinfo{person}{Nisarg Kothari}, \bibinfo{person}{Joonseok Lee}, \bibinfo{person}{Paul Natsev}, \bibinfo{person}{George Toderici}, \bibinfo{person}{Balakrishnan Varadarajan}, {and} \bibinfo{person}{Sudheendra Vijayanarasimhan}.} \bibinfo{year}{2016}\natexlab{}.
\newblock \showarticletitle{Youtube-8m: A large-scale video classification benchmark}.
\newblock \bibinfo{journal}{\emph{arXiv preprint arXiv:1609.08675}} (\bibinfo{year}{2016}).
\newblock


\bibitem[Bai et~al\mbox{.}(2025)]%
        {bai2025qwen25vltechnicalreport}
\bibfield{author}{\bibinfo{person}{Shuai Bai}, \bibinfo{person}{Keqin Chen}, \bibinfo{person}{Xuejing Liu}, \bibinfo{person}{Jialin Wang}, \bibinfo{person}{Wenbin Ge}, \bibinfo{person}{Sibo Song}, \bibinfo{person}{Kai Dang}, \bibinfo{person}{Peng Wang}, \bibinfo{person}{Shijie Wang}, \bibinfo{person}{Jun Tang}, \bibinfo{person}{Humen Zhong}, \bibinfo{person}{Yuanzhi Zhu}, \bibinfo{person}{Mingkun Yang}, \bibinfo{person}{Zhaohai Li}, \bibinfo{person}{Jianqiang Wan}, \bibinfo{person}{Pengfei Wang}, \bibinfo{person}{Wei Ding}, \bibinfo{person}{Zheren Fu}, \bibinfo{person}{Yiheng Xu}, \bibinfo{person}{Jiabo Ye}, \bibinfo{person}{Xi Zhang}, \bibinfo{person}{Tianbao Xie}, \bibinfo{person}{Zesen Cheng}, \bibinfo{person}{Hang Zhang}, \bibinfo{person}{Zhibo Yang}, \bibinfo{person}{Haiyang Xu}, {and} \bibinfo{person}{Junyang Lin}.} \bibinfo{year}{2025}\natexlab{}.
\newblock \bibinfo{title}{Qwen2.5-VL Technical Report}.
\newblock
\showeprint[arxiv]{2502.13923}~[cs.CV]
\urldef\tempurl%
\url{https://arxiv.org/abs/2502.13923}
\showURL{%
\tempurl}


\bibitem[Caba~Heilbron et~al\mbox{.}(2015)]%
        {caba2015activitynet}
\bibfield{author}{\bibinfo{person}{Fabian Caba~Heilbron}, \bibinfo{person}{Victor Escorcia}, \bibinfo{person}{Bernard Ghanem}, {and} \bibinfo{person}{Juan Carlos~Niebles}.} \bibinfo{year}{2015}\natexlab{}.
\newblock \showarticletitle{Activitynet: A large-scale video benchmark for human activity understanding}. In \bibinfo{booktitle}{\emph{Proceedings of the ieee conference on computer vision and pattern recognition}}. \bibinfo{pages}{961--970}.
\newblock


\bibitem[Chambers and Jurafsky(2008)]%
        {chambers2008unsupervised}
\bibfield{author}{\bibinfo{person}{Nathanael Chambers} {and} \bibinfo{person}{Dan Jurafsky}.} \bibinfo{year}{2008}\natexlab{}.
\newblock \showarticletitle{Unsupervised learning of narrative event chains}. In \bibinfo{booktitle}{\emph{Proceedings of ACL-08: HLT}}. \bibinfo{pages}{789--797}.
\newblock


\bibitem[Chambers and Jurafsky(2009)]%
        {chambers2009unsupervised}
\bibfield{author}{\bibinfo{person}{Nathanael Chambers} {and} \bibinfo{person}{Dan Jurafsky}.} \bibinfo{year}{2009}\natexlab{}.
\newblock \showarticletitle{Unsupervised learning of narrative schemas and their participants}. In \bibinfo{booktitle}{\emph{Proceedings of the Joint Conference of the 47th Annual Meeting of the ACL and the 4th International Joint Conference on Natural Language Processing of the AFNLP}}. \bibinfo{pages}{602--610}.
\newblock


\bibitem[Cheng et~al\mbox{.}(2025)]%
        {cheng2025tempura}
\bibfield{author}{\bibinfo{person}{Jen-Hao Cheng}, \bibinfo{person}{Vivian Wang}, \bibinfo{person}{Huayu Wang}, \bibinfo{person}{Huapeng Zhou}, \bibinfo{person}{Yi-Hao Peng}, \bibinfo{person}{Hou-I Liu}, \bibinfo{person}{Hsiang-Wei Huang}, \bibinfo{person}{Kuang-Ming Chen}, \bibinfo{person}{Cheng-Yen Yang}, \bibinfo{person}{Wenhao Chai}, {et~al\mbox{.}}} \bibinfo{year}{2025}\natexlab{}.
\newblock \showarticletitle{Tempura: Temporal event masked prediction and understanding for reasoning in action}.
\newblock \bibinfo{journal}{\emph{arXiv preprint arXiv:2505.01583}} (\bibinfo{year}{2025}).
\newblock


\bibitem[Damen et~al\mbox{.}(2020)]%
        {damen2020epic}
\bibfield{author}{\bibinfo{person}{Dima Damen}, \bibinfo{person}{Hazel Doughty}, \bibinfo{person}{Giovanni~Maria Farinella}, \bibinfo{person}{Sanja Fidler}, \bibinfo{person}{Antonino Furnari}, \bibinfo{person}{Evangelos Kazakos}, \bibinfo{person}{Davide Moltisanti}, \bibinfo{person}{Jonathan Munro}, \bibinfo{person}{Toby Perrett}, \bibinfo{person}{Will Price}, {et~al\mbox{.}}} \bibinfo{year}{2020}\natexlab{}.
\newblock \showarticletitle{The epic-kitchens dataset: Collection, challenges and baselines}.
\newblock \bibinfo{journal}{\emph{IEEE Transactions on Pattern Analysis and Machine Intelligence}} \bibinfo{volume}{43}, \bibinfo{number}{11} (\bibinfo{year}{2020}), \bibinfo{pages}{4125--4141}.
\newblock


\bibitem[Gehring et~al\mbox{.}(2017)]%
        {gehring2017convolutionalsequencesequencelearning}
\bibfield{author}{\bibinfo{person}{Jonas Gehring}, \bibinfo{person}{Michael Auli}, \bibinfo{person}{David Grangier}, \bibinfo{person}{Denis Yarats}, {and} \bibinfo{person}{Yann~N Dauphin}.} \bibinfo{year}{2017}\natexlab{}.
\newblock \showarticletitle{Convolutional sequence to sequence learning}. In \bibinfo{booktitle}{\emph{International conference on machine learning}}. PMLR, \bibinfo{pages}{1243--1252}.
\newblock


\bibitem[Girdhar and Grauman(2021)]%
        {girdhar2021anticipative}
\bibfield{author}{\bibinfo{person}{Rohit Girdhar} {and} \bibinfo{person}{Kristen Grauman}.} \bibinfo{year}{2021}\natexlab{}.
\newblock \showarticletitle{Anticipative video transformer}. In \bibinfo{booktitle}{\emph{Proceedings of the IEEE/CVF international conference on computer vision}}. \bibinfo{pages}{13505--13515}.
\newblock


\bibitem[Gkalelis et~al\mbox{.}(2022)]%
        {gkalelis2022vigat}
\bibfield{author}{\bibinfo{person}{Nikolaos Gkalelis}, \bibinfo{person}{Dimitrios Daskalakis}, {and} \bibinfo{person}{Vasileios Mezaris}.} \bibinfo{year}{2022}\natexlab{}.
\newblock \showarticletitle{ViGAT: Bottom-up event recognition and explanation in video using factorized graph attention network}.
\newblock \bibinfo{journal}{\emph{IEEE Access}}  \bibinfo{volume}{10} (\bibinfo{year}{2022}), \bibinfo{pages}{108797--108816}.
\newblock


\bibitem[Granroth-Wilding and Clark(2016)]%
        {granroth2016happens}
\bibfield{author}{\bibinfo{person}{Mark Granroth-Wilding} {and} \bibinfo{person}{Stephen Clark}.} \bibinfo{year}{2016}\natexlab{}.
\newblock \showarticletitle{What happens next? event prediction using a compositional neural network model}. In \bibinfo{booktitle}{\emph{Proceedings of the AAAI Conference on Artificial Intelligence}}, Vol.~\bibinfo{volume}{30}.
\newblock


\bibitem[Grauman et~al\mbox{.}(2022)]%
        {grauman2022ego4d}
\bibfield{author}{\bibinfo{person}{Kristen Grauman}, \bibinfo{person}{Andrew Westbury}, \bibinfo{person}{Eugene Byrne}, \bibinfo{person}{Zachary Chavis}, \bibinfo{person}{Antonino Furnari}, \bibinfo{person}{Rohit Girdhar}, \bibinfo{person}{Jackson Hamburger}, \bibinfo{person}{Hao Jiang}, \bibinfo{person}{Miao Liu}, \bibinfo{person}{Xingyu Liu}, {et~al\mbox{.}}} \bibinfo{year}{2022}\natexlab{}.
\newblock \showarticletitle{Ego4d: Around the world in 3,000 hours of egocentric video}. In \bibinfo{booktitle}{\emph{Proceedings of the IEEE/CVF conference on computer vision and pattern recognition}}. \bibinfo{pages}{18995--19012}.
\newblock


\bibitem[Grauman et~al\mbox{.}(2024)]%
        {grauman2024ego}
\bibfield{author}{\bibinfo{person}{Kristen Grauman}, \bibinfo{person}{Andrew Westbury}, \bibinfo{person}{Lorenzo Torresani}, \bibinfo{person}{Kris Kitani}, \bibinfo{person}{Jitendra Malik}, \bibinfo{person}{Triantafyllos Afouras}, \bibinfo{person}{Kumar Ashutosh}, \bibinfo{person}{Vijay Baiyya}, \bibinfo{person}{Siddhant Bansal}, \bibinfo{person}{Bikram Boote}, {et~al\mbox{.}}} \bibinfo{year}{2024}\natexlab{}.
\newblock \showarticletitle{Ego-exo4d: Understanding skilled human activity from first-and third-person perspectives}. In \bibinfo{booktitle}{\emph{Proceedings of the IEEE/CVF Conference on Computer Vision and Pattern Recognition}}. \bibinfo{pages}{19383--19400}.
\newblock


\bibitem[Gu et~al\mbox{.}(2018)]%
        {gu2018ava}
\bibfield{author}{\bibinfo{person}{Chunhui Gu}, \bibinfo{person}{Chen Sun}, \bibinfo{person}{David~A Ross}, \bibinfo{person}{Carl Vondrick}, \bibinfo{person}{Caroline Pantofaru}, \bibinfo{person}{Yeqing Li}, \bibinfo{person}{Sudheendra Vijayanarasimhan}, \bibinfo{person}{George Toderici}, \bibinfo{person}{Susanna Ricco}, \bibinfo{person}{Rahul Sukthankar}, {et~al\mbox{.}}} \bibinfo{year}{2018}\natexlab{}.
\newblock \showarticletitle{Ava: A video dataset of spatio-temporally localized atomic visual actions}. In \bibinfo{booktitle}{\emph{Proceedings of the IEEE conference on computer vision and pattern recognition}}. \bibinfo{pages}{6047--6056}.
\newblock


\bibitem[He et~al\mbox{.}(2016)]%
        {he2016deep}
\bibfield{author}{\bibinfo{person}{Kaiming He}, \bibinfo{person}{Xiangyu Zhang}, \bibinfo{person}{Shaoqing Ren}, {and} \bibinfo{person}{Jian Sun}.} \bibinfo{year}{2016}\natexlab{}.
\newblock \showarticletitle{Deep residual learning for image recognition}. In \bibinfo{booktitle}{\emph{Proceedings of the IEEE conference on computer vision and pattern recognition}}. \bibinfo{pages}{770--778}.
\newblock


\bibitem[Helff et~al\mbox{.}(2024)]%
        {helff2024vloldiagnosticdatasetvisual}
\bibfield{author}{\bibinfo{person}{Lukas Helff}, \bibinfo{person}{Wolfgang Stammer}, \bibinfo{person}{Hikaru Shindo}, \bibinfo{person}{Devendra~Singh Dhami}, {and} \bibinfo{person}{Kristian Kersting}.} \bibinfo{year}{2024}\natexlab{}.
\newblock \bibinfo{title}{V-LoL: A Diagnostic Dataset for Visual Logical Learning}.
\newblock
\showeprint[arxiv]{2306.07743}~[cs.AI]
\urldef\tempurl%
\url{https://arxiv.org/abs/2306.07743}
\showURL{%
\tempurl}


\bibitem[Jans et~al\mbox{.}(2012)]%
        {jans2012skip}
\bibfield{author}{\bibinfo{person}{Bram Jans}, \bibinfo{person}{Steven Bethard}, \bibinfo{person}{Ivan Vulic}, {and} \bibinfo{person}{Marie-Francine Moens}.} \bibinfo{year}{2012}\natexlab{}.
\newblock \showarticletitle{Skip n-grams and ranking functions for predicting script events}. In \bibinfo{booktitle}{\emph{Proceedings of the 13th Conference of the European Chapter of the Association for Computational Linguistics (EACL 2012)}}. ACL; East Stroudsburg, PA, \bibinfo{pages}{336--344}.
\newblock


\bibitem[Ji et~al\mbox{.}(2020)]%
        {ji2020action}
\bibfield{author}{\bibinfo{person}{Jingwei Ji}, \bibinfo{person}{Ranjay Krishna}, \bibinfo{person}{Li Fei-Fei}, {and} \bibinfo{person}{Juan~Carlos Niebles}.} \bibinfo{year}{2020}\natexlab{}.
\newblock \showarticletitle{Action genome: Actions as compositions of spatio-temporal scene graphs}. In \bibinfo{booktitle}{\emph{Proceedings of the IEEE/CVF conference on computer vision and pattern recognition}}. \bibinfo{pages}{10236--10247}.
\newblock


\bibitem[Kim et~al\mbox{.}(2024)]%
        {kim2024palm}
\bibfield{author}{\bibinfo{person}{Sanghwan Kim}, \bibinfo{person}{Daoji Huang}, \bibinfo{person}{Yongqin Xian}, \bibinfo{person}{Otmar Hilliges}, \bibinfo{person}{Luc Van~Gool}, {and} \bibinfo{person}{Xi Wang}.} \bibinfo{year}{2024}\natexlab{}.
\newblock \showarticletitle{Palm: Predicting actions through language models}. In \bibinfo{booktitle}{\emph{European Conference on Computer Vision}}. Springer, \bibinfo{pages}{140--158}.
\newblock


\bibitem[Kipf and Welling(2016)]%
        {kipf2016semi}
\bibfield{author}{\bibinfo{person}{Thomas~N Kipf} {and} \bibinfo{person}{Max Welling}.} \bibinfo{year}{2016}\natexlab{}.
\newblock \showarticletitle{Semi-supervised classification with graph convolutional networks}.
\newblock \bibinfo{journal}{\emph{arXiv preprint arXiv:1609.02907}} (\bibinfo{year}{2016}).
\newblock


\bibitem[Kong and Fu(2022)]%
        {kong2022human}
\bibfield{author}{\bibinfo{person}{Yu Kong} {and} \bibinfo{person}{Yun Fu}.} \bibinfo{year}{2022}\natexlab{}.
\newblock \showarticletitle{Human action recognition and prediction: A survey}.
\newblock \bibinfo{journal}{\emph{International Journal of Computer Vision}} \bibinfo{volume}{130}, \bibinfo{number}{5} (\bibinfo{year}{2022}), \bibinfo{pages}{1366--1401}.
\newblock


\bibitem[Ky and Tuyen(2018)]%
        {ky2018higher}
\bibfield{author}{\bibinfo{person}{Dao~Xuan Ky} {and} \bibinfo{person}{Luc~Tri Tuyen}.} \bibinfo{year}{2018}\natexlab{}.
\newblock \showarticletitle{A Higher order Markov model for time series forecasting}.
\newblock \bibinfo{journal}{\emph{International Journal of Applied Mathematics and Statistics}} \bibinfo{volume}{57}, \bibinfo{number}{3} (\bibinfo{year}{2018}), \bibinfo{pages}{1--18}.
\newblock


\bibitem[Lei et~al\mbox{.}(2020)]%
        {lei2020more}
\bibfield{author}{\bibinfo{person}{Jie Lei}, \bibinfo{person}{Licheng Yu}, \bibinfo{person}{Tamara Berg}, {and} \bibinfo{person}{Mohit Bansal}.} \bibinfo{year}{2020}\natexlab{}.
\newblock \showarticletitle{What is More Likely to Happen Next? Video-and-Language Future Event Prediction}. In \bibinfo{booktitle}{\emph{Proceedings of the 2020 Conference on Empirical Methods in Natural Language Processing (EMNLP)}}. \bibinfo{pages}{8769--8784}.
\newblock


\bibitem[Li and Kitani(2013)]%
        {li2013model}
\bibfield{author}{\bibinfo{person}{Cheng Li} {and} \bibinfo{person}{Kris~M Kitani}.} \bibinfo{year}{2013}\natexlab{}.
\newblock \showarticletitle{Model recommendation with virtual probes for egocentric hand detection}. In \bibinfo{booktitle}{\emph{Proceedings of the IEEE International Conference on Computer Vision}}. \bibinfo{pages}{2624--2631}.
\newblock


\bibitem[Liang et~al\mbox{.}(2025)]%
        {Liang2025videvent}
\bibfield{author}{\bibinfo{person}{Baoyu Liang}, \bibinfo{person}{Qile Su}, \bibinfo{person}{Shoutai Zhu}, \bibinfo{person}{Yuchen Liang}, {and} \bibinfo{person}{Chao Tong}.} \bibinfo{year}{2025}\natexlab{}.
\newblock \bibinfo{title}{VidEvent: A Large Dataset for Understanding Dynamic Evolution of Events in Videos}.
\newblock
\showeprint[arxiv]{2506.02448}~[cs.CV]
\urldef\tempurl%
\url{https://arxiv.org/abs/2506.02448}
\showURL{%
\tempurl}


\bibitem[Lin et~al\mbox{.}(2024)]%
        {lin2024videollavalearningunitedvisual}
\bibfield{author}{\bibinfo{person}{Bin Lin}, \bibinfo{person}{Yang Ye}, \bibinfo{person}{Bin Zhu}, \bibinfo{person}{Jiaxi Cui}, \bibinfo{person}{Munan Ning}, \bibinfo{person}{Peng Jin}, {and} \bibinfo{person}{Li Yuan}.} \bibinfo{year}{2024}\natexlab{}.
\newblock \showarticletitle{Video-LLaVA: Learning United Visual Representation by Alignment Before Projection}. In \bibinfo{booktitle}{\emph{Proceedings of the 2024 Conference on Empirical Methods in Natural Language Processing}}. \bibinfo{pages}{5971--5984}.
\newblock


\bibitem[Liu et~al\mbox{.}(2020)]%
        {liu2020forecasting}
\bibfield{author}{\bibinfo{person}{Miao Liu}, \bibinfo{person}{Siyu Tang}, \bibinfo{person}{Yin Li}, {and} \bibinfo{person}{James~M Rehg}.} \bibinfo{year}{2020}\natexlab{}.
\newblock \showarticletitle{Forecasting human-object interaction: joint prediction of motor attention and actions in first person video}. In \bibinfo{booktitle}{\emph{Computer Vision--ECCV 2020: 16th European Conference, Glasgow, UK, August 23--28, 2020, Proceedings, Part I 16}}. Springer, \bibinfo{pages}{704--721}.
\newblock


\bibitem[Liu et~al\mbox{.}(2024)]%
        {liu2024grounding}
\bibfield{author}{\bibinfo{person}{Shilong Liu}, \bibinfo{person}{Zhaoyang Zeng}, \bibinfo{person}{Tianhe Ren}, \bibinfo{person}{Feng Li}, \bibinfo{person}{Hao Zhang}, \bibinfo{person}{Jie Yang}, \bibinfo{person}{Qing Jiang}, \bibinfo{person}{Chunyuan Li}, \bibinfo{person}{Jianwei Yang}, \bibinfo{person}{Hang Su}, {et~al\mbox{.}}} \bibinfo{year}{2024}\natexlab{}.
\newblock \showarticletitle{Grounding dino: Marrying dino with grounded pre-training for open-set object detection}. In \bibinfo{booktitle}{\emph{European Conference on Computer Vision}}. Springer, \bibinfo{pages}{38--55}.
\newblock


\bibitem[Mooney and DeJong(1985)]%
        {mooney1985learning}
\bibfield{author}{\bibinfo{person}{Raymond~J Mooney} {and} \bibinfo{person}{Gerald DeJong}.} \bibinfo{year}{1985}\natexlab{}.
\newblock \showarticletitle{Learning schemata for natural language processing.}. In \bibinfo{booktitle}{\emph{Ijcai}}. \bibinfo{pages}{681--687}.
\newblock


\bibitem[Mun et~al\mbox{.}(2017)]%
        {mun2017text}
\bibfield{author}{\bibinfo{person}{Jonghwan Mun}, \bibinfo{person}{Minsu Cho}, {and} \bibinfo{person}{Bohyung Han}.} \bibinfo{year}{2017}\natexlab{}.
\newblock \showarticletitle{Text-guided attention model for image captioning}. In \bibinfo{booktitle}{\emph{Proceedings of the AAAI conference on artificial intelligence}}, Vol.~\bibinfo{volume}{31}.
\newblock


\bibitem[Nakamura et~al\mbox{.}(2021)]%
        {nakamura2021sensoraugmented}
\bibfield{author}{\bibinfo{person}{Katsuyuki Nakamura}, \bibinfo{person}{Hiroki Ohashi}, {and} \bibinfo{person}{Mitsuhiro Okada}.} \bibinfo{year}{2021}\natexlab{}.
\newblock \showarticletitle{Sensor-Augmented Egocentric-Video Captioning with Dynamic Modal Attention}. In \bibinfo{booktitle}{\emph{ACM International Conference on Multimedia (MM)}}.
\newblock


\bibitem[Persia et~al\mbox{.}(2020)]%
        {persia2020fast}
\bibfield{author}{\bibinfo{person}{Fabio Persia}, \bibinfo{person}{Daniela D'Auria}, {and} \bibinfo{person}{Giovanni Pilato}.} \bibinfo{year}{2020}\natexlab{}.
\newblock \showarticletitle{Fast learning and prediction of event sequences in a robotic system}. In \bibinfo{booktitle}{\emph{2020 Fourth IEEE International Conference on Robotic Computing (IRC)}}. IEEE, \bibinfo{pages}{447--452}.
\newblock


\bibitem[Pichotta and Mooney(2014)]%
        {pichotta2014statistical}
\bibfield{author}{\bibinfo{person}{Karl Pichotta} {and} \bibinfo{person}{Raymond Mooney}.} \bibinfo{year}{2014}\natexlab{}.
\newblock \showarticletitle{Statistical script learning with multi-argument events}. In \bibinfo{booktitle}{\emph{Proceedings of the 14th Conference of the European Chapter of the Association for Computational Linguistics}}. \bibinfo{pages}{220--229}.
\newblock


\bibitem[Qiu et~al\mbox{.}(2025)]%
        {qiu2025stepenhancingvideollmscompositional}
\bibfield{author}{\bibinfo{person}{Haiyi Qiu}, \bibinfo{person}{Minghe Gao}, \bibinfo{person}{Long Qian}, \bibinfo{person}{Kaihang Pan}, \bibinfo{person}{Qifan Yu}, \bibinfo{person}{Juncheng Li}, \bibinfo{person}{Wenjie Wang}, \bibinfo{person}{Siliang Tang}, \bibinfo{person}{Yueting Zhuang}, {and} \bibinfo{person}{Tat-Seng Chua}.} \bibinfo{year}{2025}\natexlab{}.
\newblock \bibinfo{title}{STEP: Enhancing Video-LLMs' Compositional Reasoning by Spatio-Temporal Graph-guided Self-Training}.
\newblock
\showeprint[arxiv]{2412.00161}~[cs.CV]
\urldef\tempurl%
\url{https://arxiv.org/abs/2412.00161}
\showURL{%
\tempurl}


\bibitem[Radford et~al\mbox{.}(2021)]%
        {radford2021learning}
\bibfield{author}{\bibinfo{person}{Alec Radford}, \bibinfo{person}{Jong~Wook Kim}, \bibinfo{person}{Chris Hallacy}, \bibinfo{person}{Aditya Ramesh}, \bibinfo{person}{Gabriel Goh}, \bibinfo{person}{Sandhini Agarwal}, \bibinfo{person}{Girish Sastry}, \bibinfo{person}{Amanda Askell}, \bibinfo{person}{Pamela Mishkin}, \bibinfo{person}{Jack Clark}, {et~al\mbox{.}}} \bibinfo{year}{2021}\natexlab{}.
\newblock \showarticletitle{Learning transferable visual models from natural language supervision}. In \bibinfo{booktitle}{\emph{International conference on machine learning}}. PmLR, \bibinfo{pages}{8748--8763}.
\newblock


\bibitem[Rodin et~al\mbox{.}(2024)]%
        {rodin2024action}
\bibfield{author}{\bibinfo{person}{Ivan Rodin}, \bibinfo{person}{Antonino Furnari}, \bibinfo{person}{Kyle Min}, \bibinfo{person}{Subarna Tripathi}, {and} \bibinfo{person}{Giovanni~Maria Farinella}.} \bibinfo{year}{2024}\natexlab{}.
\newblock \showarticletitle{Action scene graphs for long-form understanding of egocentric videos}. In \bibinfo{booktitle}{\emph{Proceedings of the IEEE/CVF Conference on Computer Vision and Pattern Recognition}}. \bibinfo{pages}{18622--18632}.
\newblock


\bibitem[Roy et~al\mbox{.}(2024)]%
        {roy2024interactionregionvisualtransformer}
\bibfield{author}{\bibinfo{person}{Debaditya Roy}, \bibinfo{person}{Ramanathan Rajendiran}, {and} \bibinfo{person}{Basura Fernando}.} \bibinfo{year}{2024}\natexlab{}.
\newblock \showarticletitle{Interaction region visual transformer for egocentric action anticipation}. In \bibinfo{booktitle}{\emph{Proceedings of the IEEE/CVF Winter Conference on Applications of Computer Vision}}. \bibinfo{pages}{6740--6750}.
\newblock


\bibitem[Rudinger et~al\mbox{.}(2015)]%
        {rudinger2015learning}
\bibfield{author}{\bibinfo{person}{Rachel Rudinger}, \bibinfo{person}{Vera Demberg}, \bibinfo{person}{Ashutosh Modi}, \bibinfo{person}{Benjamin Van~Durme}, {and} \bibinfo{person}{Manfred Pinkal}.} \bibinfo{year}{2015}\natexlab{}.
\newblock \showarticletitle{Learning to predict script events from domain-specific text}. In \bibinfo{booktitle}{\emph{Proceedings of the Fourth Joint Conference on Lexical and Computational Semantics}}. \bibinfo{pages}{205--210}.
\newblock


\bibitem[Rumi et~al\mbox{.}(2018)]%
        {rumi2018crime}
\bibfield{author}{\bibinfo{person}{Shakila~Khan Rumi}, \bibinfo{person}{Ke Deng}, {and} \bibinfo{person}{Flora~Dilys Salim}.} \bibinfo{year}{2018}\natexlab{}.
\newblock \showarticletitle{Crime event prediction with dynamic features}.
\newblock \bibinfo{journal}{\emph{EPJ Data Science}} \bibinfo{volume}{7}, \bibinfo{number}{1} (\bibinfo{year}{2018}), \bibinfo{pages}{43}.
\newblock


\bibitem[Sadhu et~al\mbox{.}(2021)]%
        {sadhu2021visual}
\bibfield{author}{\bibinfo{person}{Arka Sadhu}, \bibinfo{person}{Tanmay Gupta}, \bibinfo{person}{Mark Yatskar}, \bibinfo{person}{Ram Nevatia}, {and} \bibinfo{person}{Aniruddha Kembhavi}.} \bibinfo{year}{2021}\natexlab{}.
\newblock \showarticletitle{Visual semantic role labeling for video understanding}. In \bibinfo{booktitle}{\emph{Proceedings of the IEEE/CVF Conference on Computer Vision and Pattern Recognition}}. \bibinfo{pages}{5589--5600}.
\newblock


\bibitem[Schank and Abelson(1975)]%
        {schank1975scripts}
\bibfield{author}{\bibinfo{person}{Roger~C Schank} {and} \bibinfo{person}{Robert~P Abelson}.} \bibinfo{year}{1975}\natexlab{}.
\newblock \showarticletitle{Scripts, plans and goals}. In \bibinfo{booktitle}{\emph{Proceedings of the 4th International Joint Conference on Artificial Intelligence. IJCAI}}, Vol.~\bibinfo{volume}{1}.
\newblock


\bibitem[Silva et~al\mbox{.}(2018)]%
        {silva2018weighted}
\bibfield{author}{\bibinfo{person}{Michel Silva}, \bibinfo{person}{Washington Ramos}, \bibinfo{person}{Joao Ferreira}, \bibinfo{person}{Felipe Chamone}, \bibinfo{person}{Mario Campos}, {and} \bibinfo{person}{Erickson~R Nascimento}.} \bibinfo{year}{2018}\natexlab{}.
\newblock \showarticletitle{A weighted sparse sampling and smoothing frame transition approach for semantic fast-forward first-person videos}. In \bibinfo{booktitle}{\emph{Proceedings of the IEEE Conference on Computer Vision and Pattern Recognition}}. \bibinfo{pages}{2383--2392}.
\newblock


\bibitem[Su et~al\mbox{.}(2024)]%
        {su2023roformerenhancedtransformerrotary}
\bibfield{author}{\bibinfo{person}{Jianlin Su}, \bibinfo{person}{Murtadha Ahmed}, \bibinfo{person}{Yu Lu}, \bibinfo{person}{Shengfeng Pan}, \bibinfo{person}{Wen Bo}, {and} \bibinfo{person}{Yunfeng Liu}.} \bibinfo{year}{2024}\natexlab{}.
\newblock \showarticletitle{Roformer: Enhanced transformer with rotary position embedding}.
\newblock \bibinfo{journal}{\emph{Neurocomputing}}  \bibinfo{volume}{568} (\bibinfo{year}{2024}), \bibinfo{pages}{127063}.
\newblock


\bibitem[Tang et~al\mbox{.}(2019)]%
        {tang2019coin}
\bibfield{author}{\bibinfo{person}{Yansong Tang}, \bibinfo{person}{Dajun Ding}, \bibinfo{person}{Yongming Rao}, \bibinfo{person}{Yu Zheng}, \bibinfo{person}{Danyang Zhang}, \bibinfo{person}{Lili Zhao}, \bibinfo{person}{Jiwen Lu}, {and} \bibinfo{person}{Jie Zhou}.} \bibinfo{year}{2019}\natexlab{}.
\newblock \showarticletitle{Coin: A large-scale dataset for comprehensive instructional video analysis}. In \bibinfo{booktitle}{\emph{Proceedings of the IEEE/CVF Conference on Computer Vision and Pattern Recognition}}. \bibinfo{pages}{1207--1216}.
\newblock


\bibitem[Vaswani et~al\mbox{.}(2017)]%
        {vaswani2017attention}
\bibfield{author}{\bibinfo{person}{Ashish Vaswani}, \bibinfo{person}{Noam Shazeer}, \bibinfo{person}{Niki Parmar}, \bibinfo{person}{Jakob Uszkoreit}, \bibinfo{person}{Llion Jones}, \bibinfo{person}{Aidan~N Gomez}, \bibinfo{person}{{\L}ukasz Kaiser}, {and} \bibinfo{person}{Illia Polosukhin}.} \bibinfo{year}{2017}\natexlab{}.
\newblock \showarticletitle{Attention is all you need}.
\newblock \bibinfo{journal}{\emph{Advances in neural information processing systems}}  \bibinfo{volume}{30} (\bibinfo{year}{2017}).
\newblock


\bibitem[Velickovic et~al\mbox{.}(2018)]%
        {velickovic2018graph}
\bibfield{author}{\bibinfo{person}{Petar Velickovic}, \bibinfo{person}{Guillem Cucurull}, \bibinfo{person}{Arantxa Casanova}, \bibinfo{person}{Adriana Romero}, \bibinfo{person}{Pietro Lio}, {and} \bibinfo{person}{Yoshua Bengio}.} \bibinfo{year}{2018}\natexlab{}.
\newblock \showarticletitle{GRAPH ATTENTION NETWORKS}.
\newblock \bibinfo{journal}{\emph{stat}}  \bibinfo{volume}{1050} (\bibinfo{year}{2018}), \bibinfo{pages}{4}.
\newblock


\bibitem[Wang et~al\mbox{.}(2025)]%
        {wang2025fosteringvideoreasoningnextevent}
\bibfield{author}{\bibinfo{person}{Haonan Wang}, \bibinfo{person}{Hongfu Liu}, \bibinfo{person}{Xiangyan Liu}, \bibinfo{person}{Chao Du}, \bibinfo{person}{Kenji Kawaguchi}, \bibinfo{person}{Ye Wang}, {and} \bibinfo{person}{Tianyu Pang}.} \bibinfo{year}{2025}\natexlab{}.
\newblock \bibinfo{title}{Fostering Video Reasoning via Next-Event Prediction}.
\newblock
\showeprint[arxiv]{2505.22457}~[cs.CV]
\urldef\tempurl%
\url{https://arxiv.org/abs/2505.22457}
\showURL{%
\tempurl}


\bibitem[Wang et~al\mbox{.}(2021)]%
        {wang-etal-2021-coreference}
\bibfield{author}{\bibinfo{person}{Liming Wang}, \bibinfo{person}{Shengyu Feng}, \bibinfo{person}{Xudong Lin}, \bibinfo{person}{Manling Li}, \bibinfo{person}{Heng Ji}, {and} \bibinfo{person}{Shih-Fu Chang}.} \bibinfo{year}{2021}\natexlab{}.
\newblock \showarticletitle{Coreference by Appearance: Visually Grounded Event Coreference Resolution}. In \bibinfo{booktitle}{\emph{Proceedings of the Fourth Workshop on Computational Models of Reference, Anaphora and Coreference}}, \bibfield{editor}{\bibinfo{person}{Maciej Ogrodniczuk}, \bibinfo{person}{Sameer Pradhan}, \bibinfo{person}{Massimo Poesio}, \bibinfo{person}{Yulia Grishina}, {and} \bibinfo{person}{Vincent Ng}} (Eds.). \bibinfo{publisher}{Association for Computational Linguistics}, \bibinfo{address}{Punta Cana, Dominican Republic}, \bibinfo{pages}{132--140}.
\newblock
\href{https://doi.org/10.18653/v1/2021.crac-1.14}{doi:\nolinkurl{10.18653/v1/2021.crac-1.14}}


\bibitem[Wang et~al\mbox{.}(2019b)]%
        {wang2019dgl}
\bibfield{author}{\bibinfo{person}{Minjie Wang}, \bibinfo{person}{Da Zheng}, \bibinfo{person}{Zihao Ye}, \bibinfo{person}{Quan Gan}, \bibinfo{person}{Mufei Li}, \bibinfo{person}{Xiang Song}, \bibinfo{person}{Jinjing Zhou}, \bibinfo{person}{Chao Ma}, \bibinfo{person}{Lingfan Yu}, \bibinfo{person}{Yu Gai}, \bibinfo{person}{Tianjun Xiao}, \bibinfo{person}{Tong He}, \bibinfo{person}{George Karypis}, \bibinfo{person}{Jinyang Li}, {and} \bibinfo{person}{Zheng Zhang}.} \bibinfo{year}{2019}\natexlab{b}.
\newblock \showarticletitle{Deep Graph Library: A Graph-Centric, Highly-Performant Package for Graph Neural Networks}.
\newblock \bibinfo{journal}{\emph{arXiv preprint arXiv:1909.01315}} (\bibinfo{year}{2019}).
\newblock


\bibitem[Wang et~al\mbox{.}(2023)]%
        {wang2023caption}
\bibfield{author}{\bibinfo{person}{Teng Wang}, \bibinfo{person}{Jinrui Zhang}, \bibinfo{person}{Junjie Fei}, \bibinfo{person}{Hao Zheng}, \bibinfo{person}{Yunlong Tang}, \bibinfo{person}{Zhe Li}, \bibinfo{person}{Mingqi Gao}, {and} \bibinfo{person}{Shanshan Zhao}.} \bibinfo{year}{2023}\natexlab{}.
\newblock \showarticletitle{Caption anything: Interactive image description with diverse multimodal controls}.
\newblock \bibinfo{journal}{\emph{arXiv preprint arXiv:2305.02677}} (\bibinfo{year}{2023}).
\newblock


\bibitem[Wang et~al\mbox{.}(2019a)]%
        {wang2019vatex}
\bibfield{author}{\bibinfo{person}{Xin Wang}, \bibinfo{person}{Jiawei Wu}, \bibinfo{person}{Junkun Chen}, \bibinfo{person}{Lei Li}, \bibinfo{person}{Yuan-Fang Wang}, {and} \bibinfo{person}{William~Yang Wang}.} \bibinfo{year}{2019}\natexlab{a}.
\newblock \showarticletitle{Vatex: A large-scale, high-quality multilingual dataset for video-and-language research}. In \bibinfo{booktitle}{\emph{Proceedings of the IEEE/CVF international conference on computer vision}}. \bibinfo{pages}{4581--4591}.
\newblock


\bibitem[Xu et~al\mbox{.}(2018)]%
        {xu2018powerful}
\bibfield{author}{\bibinfo{person}{Keyulu Xu}, \bibinfo{person}{Weihua Hu}, \bibinfo{person}{Jure Leskovec}, {and} \bibinfo{person}{Stefanie Jegelka}.} \bibinfo{year}{2018}\natexlab{}.
\newblock \showarticletitle{How powerful are graph neural networks?}
\newblock \bibinfo{journal}{\emph{arXiv preprint arXiv:1810.00826}} (\bibinfo{year}{2018}).
\newblock


\bibitem[Yang et~al\mbox{.}(2020)]%
        {yang2020nargnn}
\bibfield{author}{\bibinfo{person}{Shuang Yang}, \bibinfo{person}{Fali Wang}, \bibinfo{person}{Daren Zha}, \bibinfo{person}{Cong Xue}, {and} \bibinfo{person}{Zhihao Tang}.} \bibinfo{year}{2020}\natexlab{}.
\newblock \showarticletitle{NarGNN: Narrative graph neural networks for new script event prediction problem}. In \bibinfo{booktitle}{\emph{2020 IEEE Intl Conf on Parallel \& Distributed Processing with Applications, Big Data \& Cloud Computing, Sustainable Computing \& Communications, Social Computing \& Networking (ISPA/BDCloud/SocialCom/SustainCom)}}. IEEE, \bibinfo{pages}{481--488}.
\newblock


\bibitem[Zhang et~al\mbox{.}(2020)]%
        {zhang2020does}
\bibfield{author}{\bibinfo{person}{Zhu Zhang}, \bibinfo{person}{Zhou Zhao}, \bibinfo{person}{Yang Zhao}, \bibinfo{person}{Qi Wang}, \bibinfo{person}{Huasheng Liu}, {and} \bibinfo{person}{Lianli Gao}.} \bibinfo{year}{2020}\natexlab{}.
\newblock \showarticletitle{Where Does It Exist: Spatio-Temporal Video Grounding for Multi-Form Sentences}. In \bibinfo{booktitle}{\emph{CVPR}}.
\newblock


\bibitem[Zhao and Wildes(2021)]%
        {zhao2021reviewvideopredictiveunderstanding}
\bibfield{author}{\bibinfo{person}{He Zhao} {and} \bibinfo{person}{Richard~P. Wildes}.} \bibinfo{year}{2021}\natexlab{}.
\newblock \bibinfo{title}{Review of Video Predictive Understanding: Early Action Recognition and Future Action Prediction}.
\newblock
\showeprint[arxiv]{2107.05140}~[cs.CV]
\urldef\tempurl%
\url{https://arxiv.org/abs/2107.05140}
\showURL{%
\tempurl}


\bibitem[Zhao et~al\mbox{.}(2019)]%
        {zhao2019hacs}
\bibfield{author}{\bibinfo{person}{Hang Zhao}, \bibinfo{person}{Zhicheng Yan}, \bibinfo{person}{Lorenzo Torresani}, {and} \bibinfo{person}{Antonio Torralba}.} \bibinfo{year}{2019}\natexlab{}.
\newblock \showarticletitle{HACS: Human Action Clips and Segments Dataset for Recognition and Temporal Localization}.
\newblock \bibinfo{journal}{\emph{arXiv preprint arXiv:1712.09374}} (\bibinfo{year}{2019}).
\newblock


\bibitem[Zhao(2021)]%
        {zhao2021event}
\bibfield{author}{\bibinfo{person}{Liang Zhao}.} \bibinfo{year}{2021}\natexlab{}.
\newblock \showarticletitle{Event prediction in the big data era: A systematic survey}.
\newblock \bibinfo{journal}{\emph{ACM Computing Surveys (CSUR)}} \bibinfo{volume}{54}, \bibinfo{number}{5} (\bibinfo{year}{2021}), \bibinfo{pages}{1--37}.
\newblock


\bibitem[Zheng et~al\mbox{.}(2020)]%
        {zheng2020heterogeneous}
\bibfield{author}{\bibinfo{person}{Jianming Zheng}, \bibinfo{person}{Fei Cai}, \bibinfo{person}{Yanxiang Ling}, {and} \bibinfo{person}{Honghui Chen}.} \bibinfo{year}{2020}\natexlab{}.
\newblock \showarticletitle{Heterogeneous graph neural networks to predict what happen next}. In \bibinfo{booktitle}{\emph{Proceedings of the 28th International Conference on Computational Linguistics}}. \bibinfo{pages}{328--338}.
\newblock


\end{thebibliography}

\appendix

\section{Dataset}
\label{sec:apA}

\begin{table*}[h]
    \centering
    \caption{A non-exhaustive summary on the related video datasets.}
    \label{tab:datasets}
    \begin{tabular}{c c c c c}
        \toprule
        \multirow{2}{*}{\textbf{Perspective}} & \multirow{2}{*}{\textbf{Dataset}} & \multicolumn{2}{c}{\textbf{Event Structure}} & \multirow{2}{*}{\textbf{Event Prediction Surpport}} \\ \cline{3-4}
        &  & trigger & arguments & \\ \hline
        \multirow{12}{*}{Third Person}
        & ActivityNet(2015) \cite{caba2015activitynet} & \ding{51} & \ding{55} & \ding{55} \\ 
        & Youtube8M(2016) \cite{abu2016youtube} & \ding{55} & \ding{55} & \ding{55}\\ 
        & AVA(2018) \cite{gu2018ava} & \ding{51} & \ding{55} & \ding{55}\\
        & HACS(2019) \cite{zhao2019hacs} & \ding{51} & \ding{55} & \ding{55}\\
        & COIN(2019) \cite{tang2019coin} & \ding{55} & \ding{55} & \ding{55}\\
        & Vatex(2020) \cite{wang2019vatex} & \ding{51} & \ding{51} & \ding{55}\\
        & VidSTG(2020) \cite{zhang2020does} & \ding{51} & \ding{51} & \ding{55}\\
        & VLEP(2020) \cite{lei2020more} & \ding{51} & \ding{51} & \ding{55}\\
        & VidSitu(2021) \cite{sadhu2021visual} & \ding{51} & \ding{51} & \ding{51}(only after processing)\\
        & VidEvent(2025) \cite{Liang2025videvent} & \ding{51} & \ding{51} & \ding{51}\\ \hline
        \multirow{6}{*}{First Person}
        & EgoAction(2013) \cite{li2013model} & \ding{51} & \ding{55} & \ding{55}\\
        & DoMSEV(2018) \cite{silva2018weighted} & \ding{51} & \ding{51} & \ding{55}\\
        & Epic-Kitchens(2020) \cite{damen2020epic} & \ding{51} & \ding{51} & \ding{51}(only after processing)\\
        & MMAC Captions(2021) \cite{nakamura2021sensoraugmented} & \ding{51} & \ding{51} & \ding{55}\\
        & Ego4D(2022) \cite{grauman2022ego4d} & \ding{51} & \ding{51} & \ding{51}(only after processing)\\
        & Ego-Exo4D(2024) \cite{grauman2024ego} & \ding{51} & \ding{51} & \ding{51}(only after processing)\\
        \bottomrule
        
    \end{tabular}
\end{table*}

\subsection{Dataset Composition}
Here, we provide a detailed description of all the datasets utilized in the expansion process to construct the AVEP dataset. \\
\textbf{EPIC-KITCHENS-100 \cite{damen2020epic}.} EPIC-KITCHENS-100 is a collection of $100$ hours, $20M$ frames, $90K$ actions in $700$ variable-length videos, capturing long-term unscripted activities in $45$ environments, using head-mounted cameras, which initially builed for the STA task. \\
\textbf{VidSitu \cite{sadhu2021visual}.} VidSitu, a large-scale video understanding dataset comprising $29K$ richly annotated 10-second movie clips, structures videos as a series of interconnected events, with each event containing a verb and multiple entities assigned to specific semantic roles. Clips in VidSitu are drawn from a large collection of movies $(\sim3K)$ and have been chosen to be both complex ($\sim4.2$ unique verbs within a video) as well as diverse ($\sim200$ verbs have more than $100$ annotations each). \\
\textbf{VidEvent \cite{Liang2025videvent}. } VidEvent is a large-scale dataset containing over $23,000$ well-labeled events, featuring detailed event structures, broad hierarchies, and logical relations extracted from movie recap videos, which is initially builed for video understanding. 

\subsection{Construction Procedure}
We further describe the detailed overview of the annotation process in this subsection. The entire AVEP dataset construction process can be divided into three main stages: data collection, integration and expansion, and manual verification. 

We conducted a comprehensive survey of existing event-related video datasets and categorized them based on first-person and third-person perspectives as shown in Table \ref{tab:datasets}. We evaluated the completeness of event structures based on whether they satisfy the requirements of event triggers and arguments. Datasets lacking annotated actions or containing only a limited set of action categories were considered insufficient for meeting the event trigger criteria. Similarly, datasets that failed to annotate action-related nouns or had a restricted variety of noun annotations were deemed inadequate for fulfilling the event argument requirements. Building upon the datasets that meet the event structure requirements, we further evaluated their suitability for the AVEP task. AVEP requires each sample in the dataset to contain at least three events, with clearly defined event boundaries and explicit temporal relationships between events. 

Considering both the dataset's support for the task and its scale, we selected three datasets, VidSitu, VidEvent and Epic-Kichens as the foundation datasets. We then expanded them to construct a dataset that effectively supports the AVEP task following the pipeline detailed in the subsequent sections.

\begin{figure*}[ht]
    \centering
    \includegraphics[width=0.8\linewidth]{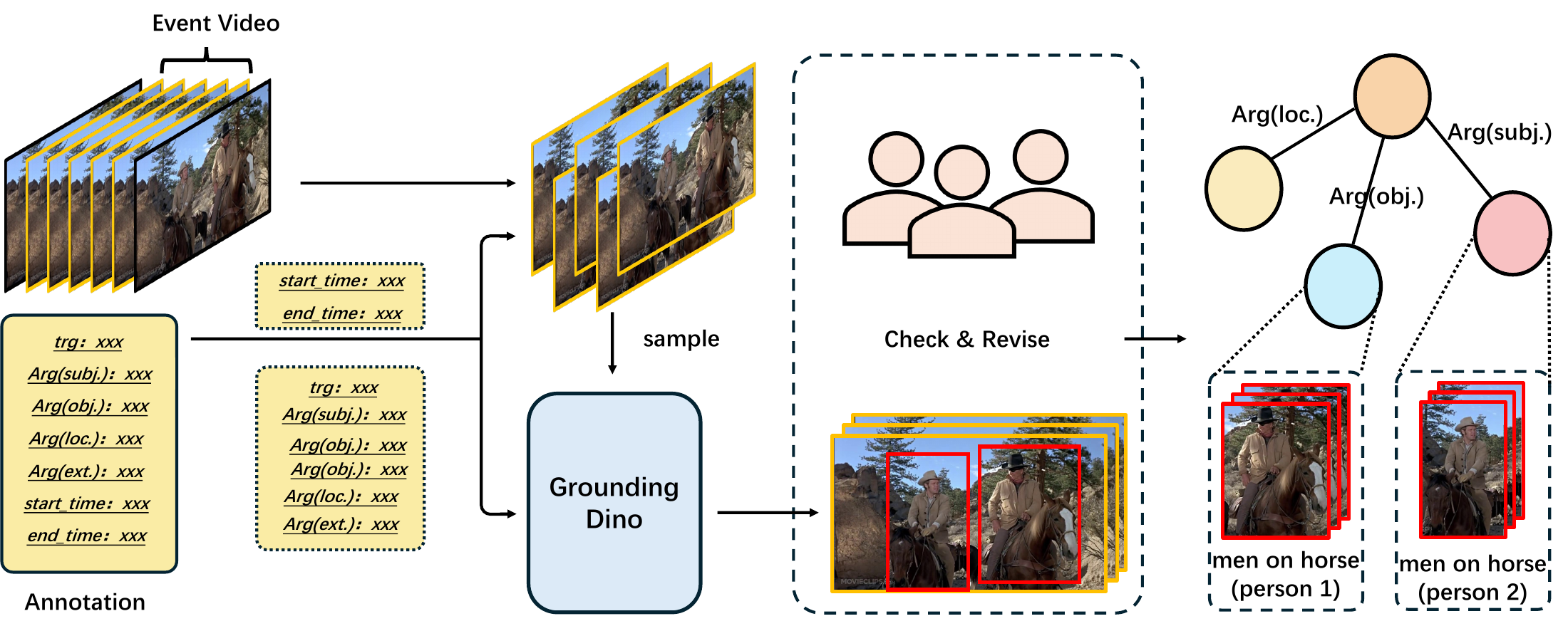}
    \caption{The annotation pipeline of AVEP. We employ the pre-trained Grounding Dino model to localize the corresponding regions in each frame based on textual annotations. Subsequently, three annotators manually verify and refine the bounding boxes according to the textual descriptions. Finally, the selected regions are cropped to generate a sequence of images, which serve as the visual representations of event graph nodes.}
    \label{fig:at_procedure}
\end{figure*}

The video datasets are composed of continuous video event segments, which are derived by partitioning the entire video, along with corresponding textual annotations of event arguments or sentences. As shown in Figure \ref{fig:at_procedure}, we can divide the entire annotation process into two parts: automatic program annotation and manual review annotation. First, we manually organized the text annotations of all datasets into a format based on event arguments. For the unannotated data with single-sentence descriptions, we utilized an argument segmentation model to automatically extract the event arguments. The event boundaries in the annotations were then used to segment the videos into a series of events. Next, we employ the DGINO model \cite{liu2024grounding} to sequentially label the bounding box corresponding to each argument in every frames of an event video clip, obtaining the visual representation for each argument. After that, three professional annotators conduct a review and correction of the automated bounding box annotations for each argument.

Three annotators spent five months correcting the data, with an average rate of around 600 samples per week. On average, each video clip required approximately 1-2 minutes to complete the full annotations. Additionally, around 35\% of the initially generated bounding boxes required manual correction to ensure accuracy. After the correction phase, an additional month was spent to cross-validation through random sampling.

Along with the argument roles, text argument annotations, and video argument annotations , we treat the text and video argument annotations as node information, while the argument roles serve as the edge relationships between the nodes, thus constructing a multimodal graph representation of the events. By arranging all the events in the video sequentially, we finally obtain the video event graph chain.

\begin{figure*}[ht]
    \centering
    \includegraphics[width=1.0\linewidth]{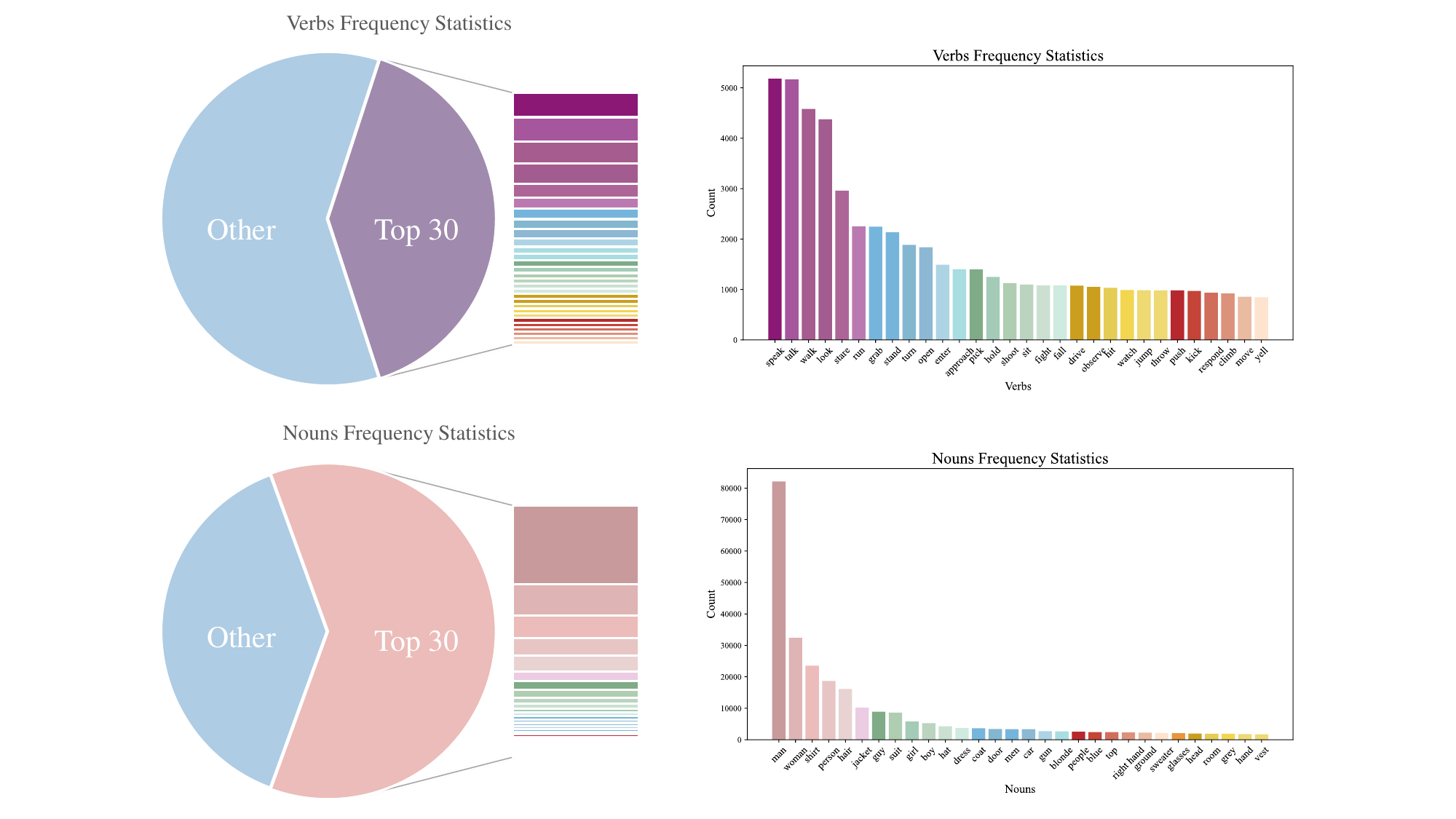}
    \caption{The data statistics on verbs and nouns in the AVEP dataset. Given the large number of verb categories ($2,284$ in total), we analyzed the distribution by calculating the proportion of the top 30 most frequent verbs versus all others, and also reported the frequency of each verb within the top 30. A similar analysis was conducted for the nouns in the dataset.}
    \label{fig:statistics}
\end{figure*}

\subsection{Data Statistics}

\begin{table}[t]
    \centering
    \caption{Statistics on splits of AVEP dataset.}
    \label{tab:count}
    \resizebox{0.8\linewidth}{!}{
    \begin{tabular}{c | c c c | c}
        \toprule
        & train & Val & Test & Total \\ \midrule
        Videos & 29793 & 1326& 4145& 35264 \\ 
        Events & 154022 & 6615 & 17771 & 178208 \\ 
        Tri.\&Args. & 434758 & 19985 & 43382 & 498125 \\ 
        %Verbs & - & - & - & 2284 \\ 
        %Nouns & - & - & - & 6927 \\ 
        \bottomrule
    \end{tabular}
    }
\end{table}

AVEP is a dataset that contains more than $178K$ video event graphs of complex structures, rich hierarchy, and logical evolutionary. To validate these characteristics, we performed a statistical analysis of the dataset from multiple perspectives, as shown in Figure \ref{fig:statistics}.

The dataset comprises a total of $35K$ video clips, with an average video length of 15 seconds. Each clip contains historical segments averaging 12 seconds in duration, followed by future events. To ensure temporal independence between events, we maintain a minimum gap of 20 frames between the last historical event and subsequent future events, effectively preventing any potential frame leakage in our dataset construction.

The dataset comprises a total of $2,284$ unique verbs. Our analysis of the verb distribution reveals a relatively balanced spread. Notably, the top 30 most frequently occurring verbs collectively constitute less than 40\% of the entire set, with no single verb overwhelmingly dominating the distribution. This level of lexical diversity ensures comprehensive coverage of the vast majority of events encountered in real-world applications.

To evaluate whether the dataset sufficiently captures the rich semantics of events, we analyzed the distribution of nouns in the annotations of $Args$. The results indicate that the dataset contains over 6,000 unique nouns. After excluding commonly used terms such as "man" and "woman," which frequently appear in annotations referring to people, we found that the remaining nouns are relatively evenly distributed. This balanced distribution ensures that our dataset aligns with the semantic richness observed in real-world applications. Additionally, we analyzed the total number of $Args$ present in each video event graph. The results show that the average of the event graphs contain  $2.8$ arguments. 

\subsection{Impact of Bounding Box Quality}
To provide a higher-quality dataset, we manually refined the automatically generated bounding boxes. Although this process reduced noise and improved overall data quality, we remain concerned about how bounding box quality impacts model performance in this task. To investigate this, we conducted experiments using the original, uncorrected bounding boxes. The results are presented in Table \ref{tab:bb_effect}.

The results show that using the original, uncorrected data indeed leads to a performance drop. Notably, the impact on noun prediction is more significant than that on verbs, which aligns with our intuition. Overall, however, the decline across all evaluation metrics remains under 10\%, indicating that the model can still be effectively trained using the original data. This also demonstrates the robustness of our method to noise in the bounding boxes.

\begin{table*}[t]
\centering
\caption{The impact of bounding box quality on AVEP task.}
\label{tab:bb_effect}
\begin{tabular}{c|cc|ccc|ccc}
\toprule[1pt]
\multirow{2}{*}{Set} & \multicolumn{2}{c|}{Verb} & \multicolumn{3}{c|}{Noun} & \multicolumn{3}{c}{Verb-Noun}\\ \cline{2-9} 
& Top1 & Top5 & P & R & F1 & P & R & F1\\ 
\midrule
val & 22.13(-0.19) & 44.93(-0.23) & 31.29(-1.81) & 42.98(-1.92) & 36.22(-1.89) & 7.40(-0.27) & 10.03(-0.85) & 8.59(-0.41) \\
test & 21.45(-1.26) & 44.32(-1.11) & 40.96(-1.76) & 48.35(-1.89) & 44.35(-1.89) & 7.32(-0.35) & 7.69(-0.02) & 7.50(-0.19) \\
\bottomrule[1pt]
\end{tabular}
\end{table*}

\section{Task Comparison}
\label{sec:apB}
To better position our task within the landscape of event reasoning and prediction, we compare it with related tasks along several key dimensions, as shown in Table \ref{tab:tasks}:
\begin{itemize}
    \item \textbf{Modelities.} Our task involves both video and text inputs, making it a multimodal task. This contrasts with many existing approaches that focus on either visual or textual input alone.
    \item \textbf{Form of historical context.} nstead of using raw video clips, we employ structured event chains to represent historical context. This structure provides richer and more explicit logical relationships between events, thereby imposing higher demands on the model’s reasoning capabilities. Moreover, representing event chains as event graph chains enables finer-grained modeling of internal event structures.
    \item \textbf{Granularity.} Events in our task are defined at a higher semantic level than low-level actions. Unlike simple action units, events encapsulate complex interactions and relationships, making the task more challenging and semantically rich.
    \item \textbf{Task formulation.} Many prior works adopt a multiple-choice format to ensure determinism and simplify evaluation. However, such formats deviate from real-world application scenarios. In contrast, we formulate our task as an open classification problem, and supplement it with human evaluation to provide a more realistic and credible measure of model performance.
\end{itemize}

\begin{table*}[h]
    \centering
    \caption{A non-exhaustive summary on the related prediction tasks.}
    \label{tab:tasks}
    \begin{tabular}{c c c c c}
        \toprule
        \textbf{Task} & \textbf{Modalities} & \textbf{Form of historical context} & \textbf{Granularity} & \textbf{Task formulation} \\ \hline
        MCNC & text & event chain & event & Multiple Choice \\
        VLEP & video & video clip & event & Multiple Choice \\
        VidEvent & text + video & event chain & event & Multiple Choice \\
        NEP & video & video clip & event & Multiple Choice \\
        Next Action Anticipation & video & frames & action & Word-based Classification \\
        LTA & video & video clip & action & Word-based Classification \\
        STA & video & video clip & action & Detection\&Classification\&Regression \\
        AVEP(ours) & text + video & event graph chain & event & Word-based Classification \\
        \bottomrule
    \end{tabular}
\end{table*}

\section{A Node-graph Hierarchical Attention Mechanism Example}
\label{sec:apC}
\begin{figure*}
    \centering
    \includegraphics[width=\linewidth]{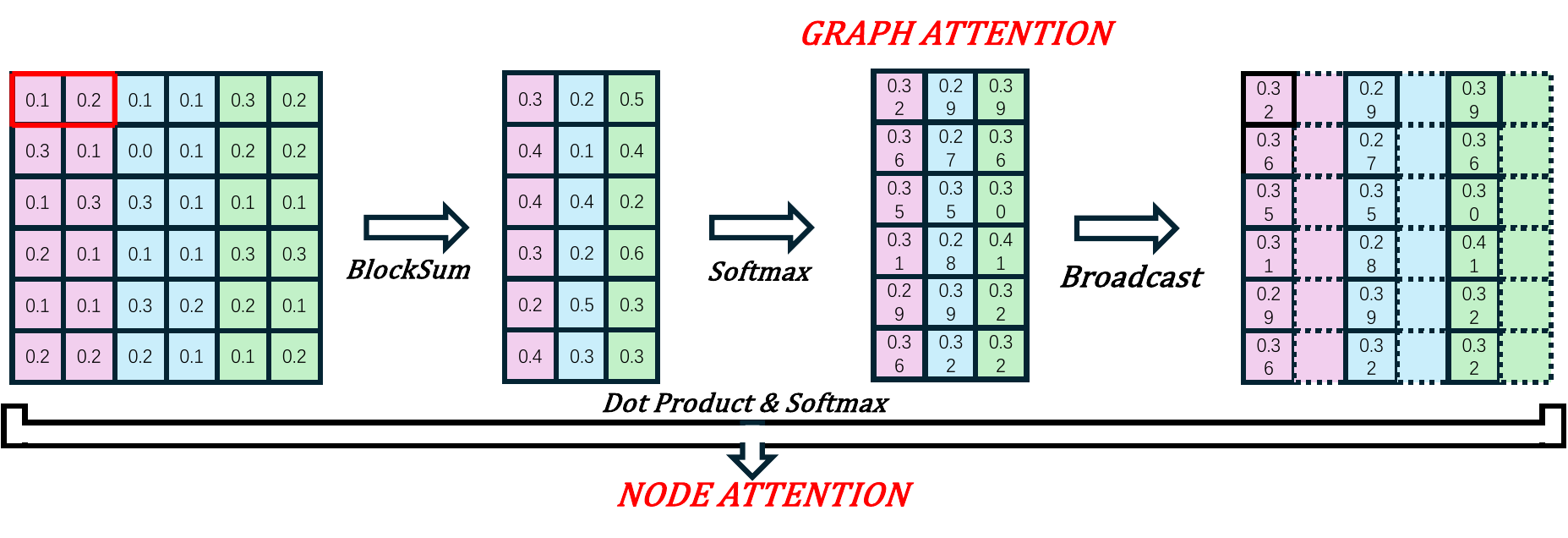}
    \caption{An example illustrating how the attention mechanism is calculated.}
    \label{pic:example}
\end{figure*}

To provide a clearer explanation of this attention mechanism, we illustrate the attention computation process using a sequence of three event graphs, each consisting of two nodes, as shown in Figure \ref{pic:example}. Different colors are used to represent different event graphs.

In the attention matrix, the red box indicates the attention scores between node pairs across different event graphs. To derive a graph-level attention score, we sum the scores within each matrix block, producing a $6\times3$ event graph-level attention matrix. By applying a softmax function in the dimension of graphs, we obtain the attention scores on the nodes for each event graph in the sequence, \textit{GRAPH ATTENTION}.

Furthermore, to integrate graph and node attention, we broadcast the computed \textit{GRAPH ATTENTION} back to their original shape $6\times6$ and perform an element-wise multiplication with the attention matrix. This operation yields the final \textit{NODE ATTENTION}, capturing both intra- and inter-event dependencies effectively.

\section{Human Evaluation}
\label{sec:apD}
Considering the inherent ambiguity in the AVEP task, we provide human evaluation results as a credible reference point for assessing model performance. To this end, we developed a dedicated web interface and invited several domain experts to participate in the evaluation. Each participant was asked to watch the video segment corresponding to the historical events of a given sample. To prevent data leakage, we trimmed the videos to exclude any frames related to future events. Based on the observed video content, the evaluators were then asked to predict the arguments of the future event. To ensure fair and meaningful evaluation, we provided a predefined vocabulary and reference options, minimizing the influence of lexical variation on prediction quality. On average, each sample took about 30 seconds to complete, and the entire human evaluation process spanned approximately two weeks.

\section{Baselines}
\label{sec:apE}
\textbf{VidEvent}\cite{Liang2025videvent} VidEvent is an event prediction model, which is proposed to predict events more likely to happen in five candidates. VidEvent is composed of two braches of vision and text. Each branch contains an encoder for video or text to extract the corresponding features, followed by a transformer encoder to make inner interactions among events within one modal. Through this two-branch architecture, VidEvent presents a strong capability in comprehensing event chains, thus achieving extraordinary results in Script event induction task.
\\\textbf{InAViT}\cite{roy2024interactionregionvisualtransformer} InAViT is a Transformer variant to model interactions by computing the change in the appearance of objects and human hands due to the execution of the actions and use those changes to refine the video representation. Specifically, it models interactions between hands and objects using Spatial Cross-Attention (SCA) and further infuses contextual information using Trajectory Cross-Attention to obtain environment-refined interaction tokens. Through these, InAViT achieves the SOTA performance in STA task.
\\\textbf{Video-LlaVA}\cite{lin2024videollavalearningunitedvisual} In this approach, the authors unify visual representation into the language feature space to advance the foundational LLM towards a unified LVLM. As a result, they establish a simple but robust LVLM baseline, Video-LLaVA, which learns from a mixed dataset of images and videos, mutually enhancing each other. Video-LLaVA achieves superior performances on a broad range of 9 image benchmarks across 5 image question-answering datasets and 4 image benchmark toolkits.
\\\textbf{Qwen2.5-VL}\cite{bai2025qwen25vltechnicalreport} Qwen2.5-VL is a comprehensive series of large language models (LLMs) designed to meet diverse needs. Compared to previous iterations, Qwen2.5 has been significantly improved during both the pre-training and post-training stages. In terms of pre-training, the authors have scaled the high-quality pre-training datasets from the previous 7 trillion tokens to 18 trillion tokens. This provides a strong foundation for common sense, expert knowledge, and reasoning capabilities. In terms of post-training, they implement intricate supervised finetuning with over 1 million samples, as well as multistage reinforcement learning. This model currently represents the state-of-the-art (SOTA) in video reasoning tasks.

\section{LVLM Experiment details}
\label{sec:apF}
We compared the performance of state-of-the-art Large Vision-Language Models (LVLMs) in video understanding tasks on the AVEP task under two different parameter scales 7B and 72B. Below, we provide detailed descriptions of the experimental setup and results.\\
\textbf{Prompts.} 
\begin{itemize}
    \item 
    $X_{system}$: \textit{"As an event prediction expert, please assist me in completing a video event prediction task."} 
    \item
    $X_0$: \textit{"I will provide you with a video segment that describes four sequential events and a list of all characters appearing in the video. Your task is to predict the future event that may occur. Please select the top five most probable verbs from the provided list of verbs and select the relevant argument(s) from the provided list of arguments."} 
    \item 
    $X_1$: \textit{"Here is the video segment describing four events." @Video} 
    \item 
    $X_2$: \textit{"Here is the verb list, please choose verbs from it. Answer with the selected verbs without output any other words." @Verb Vocabulary} 
    \item 
    $X_3$: \textit{"Here is the argument list, please select the relevant argument(s) from the provided list. Answer with the selected arguments separated with "|" without output any other words."  @ Argument List}
\end{itemize} 

\begin{figure*}[ht]
    \centering
    \includegraphics[width=\linewidth]{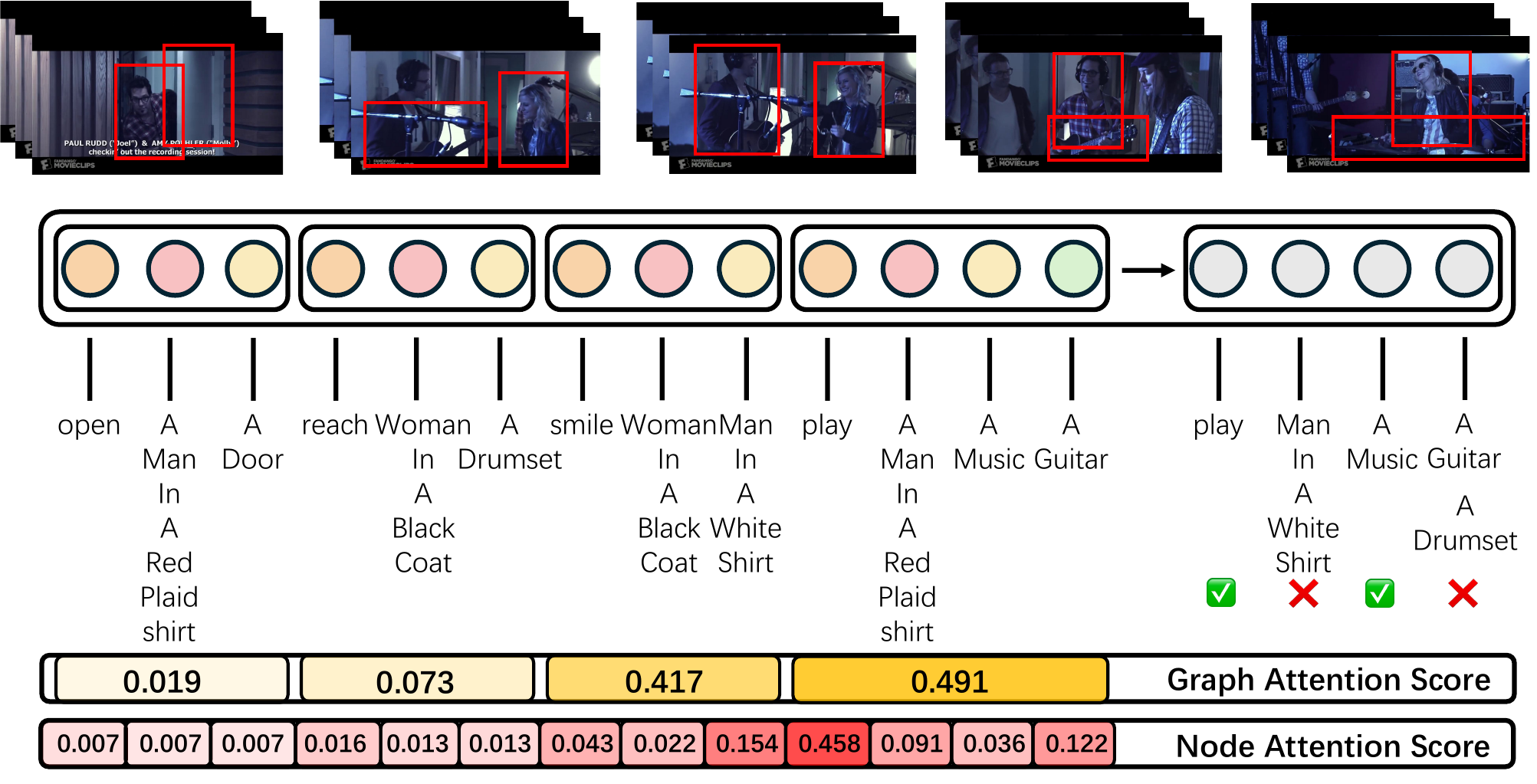}
    \caption{An example illustrating the node-graph hierarchical attention scores and the event prediction result. }
    \label{pic:result_example.pdf}
\end{figure*}

\textbf{An Example.} 

\textbf{System:} As an event prediction expert, please assist me in completing a video event prediction task. 

\textbf{User:} I will provide you with a video segment that describes four sequential events and a list of all characters appearing in the video. Your task is to predict the future event that may occur. Please select the top five most probable verbs from the provided list of verbs and select the relevant argument(s) from the provided list of arguments. 

\textbf{Assistant:} Sure, I can help you with that. Please provide me with the video segment and the list of arguments and verbs. 

\textbf{User:} Here is the video segment describing four events. @Video 

\textbf{Assistant:} Based on the video segment, here are the four events that have occurred: 1. A woman is walking down a hallway. 2. The woman enters a bathroom. 3. Another woman is in the bathroom holding a pregnancy test. 4. The first woman takes the pregnancy test from the second woman. Given these events, here are the top five most probable verbs and relevant characters for the future event: 1. Verb: "react" - Arguments: Second woman (the one who was holding the pregnancy test)  - The second woman may react to the results of the pregnancy test, either positively or negatively. 2. Verb: "talk" - Arguments: Both women - The two women may talk about the results of the pregnancy test and what it means for their future. 3. Verb: "cry" - Arguments: Either woman - Depending on the results of the pregnancy test, either woman may cry due to happiness, sadness, or shock. 4. Verb: "hug" - Arguments: Both women - The two women may hug each other in support or celebration after receiving the results of the pregnancy test. 5. Verb: "leave" - Arguments: First woman - The first woman may leave the bathroom after taking the pregnancy test from the second woman, possibly to give her some privacy or to process the information.

\textbf{User:} Here is the verb list, please choose verbs from it. Answer with the selected verbs without output any other words." @Verb Vocabulary 

\textbf{Assistant:} react talk cry hug leave 

\textbf{User:} Here is the argument list, please select the relevant argument(s) from the provided list. Answer with the selected arguments separated with "|" without output any other words."  @ Argument List 

\textbf{Assistant:} woman with brown hair | cwoman with blonde hair

\section{Frequent Questions and Answers}
\label{sec:apG}
This section addresses frequently asked questions that may arise from our main paper and experimental setup. We hope this QFA helps clarify certain design choices and implementation details.

\textbf{Q1: How dose AVEP differs from VLEP and VidEvent?}

\textbf{A1:} VLEP is an outstanding contribution to video event prediction, as it is the first to introduce a multiple-choice formulation for this task. It focuses on verb prediction by providing the subject of the event directly, which allows for a more targeted evaluation of a model’s understanding of event dynamics and its ability to anticipate future developments. However, one limitation of VLEP is that it considers only a single historical event as the context for predicting a future event. In real-world scenarios, future events are often shaped by a sequence of preceding events rather than by an isolated instance. Therefore, modeling event prediction based on event chains can offer a more realistic and informative setting. VidEvent is also an excellent work in the field of video understanding, which first proposes a comprehensive video understanding framework composed of several sub-tasks, including event prediction. However, event prediction is not the main focus of this work, and the formulation simply borrows the design of the MCNC task. This multiple-choice format, while convenient, falls short in rigorously evaluating a model’s true understanding of events. As shown in their experimental results, models tend to choose options with higher overlap in arguments with the historical context, rather than demonstrating genuine reasoning capabilities. In contrast, our proposed AVEP task requires the model to accurately predict all arguments of the future event, which forces it to learn the underlying logical relationships between events rather than relying on shallow similarity-based cues. That said, VLEP and VidEvent have made a significant contribution to the field by laying foundational insights and offering a valuable perspective for subsequent research.

\textbf{Q2: The parameter size of the SOTA model EventFormer in the experimental results is $7.3M$. Why was increasing the model's parameter size not considered to achieve better experimental results?}

\textbf{A2:} In our method, we proposed a node-graph hierarchical attention mechanism that significantly improved the model’s performance on the task. However, since the current model is primarily designed to validate the effectiveness of our proposed approach, the implementation of the attention mechanism still contains some computational redundancies that could be optimized. Therefore, increasing the model size would have negatively impacted computational efficiency. We made a deliberate trade-off between performance and efficiency. However, we are willing to develop a more optimized implementation of the attention mechanism in the future to further improve the model’s efficiency. Notably, our $7.3M$ model already outperforms other larger-scale SOTA models, which demonstrates that our proposed method can more effectively capture complex event relationships and perform event prediction. We look forward to exploring the model’s full potential in future work.

\textbf{Q3: Why did not explore more complex GNN architectures to test whether they could further improve the model's performance?}

\textbf{A3:} As shown in the experimental results, different GNN architectures can indeed influence the overall performance of the model. However, it's important to note that in our approach, the GNN module is mainly used to replace Linear later as a module for computing $QKV$ in attention, rather than serving as the core of the model. Therefore, there is no strong justification for significantly increasing the complexity of this component. 
Meanwhile, considering the scale of the event graph, we suppose that the basic GNNs possess the capability in performing effective graph embedding. 
Moreover, we suggest that increasing the model’s depth by adding more layers can be a more effective way to enhance its capacity and performance, compared to simply using more complex GNNs. However, our work only lays a solid foundation for this research direction, and we warmly welcome further exploration and improvements that may unlock more potential from the model.

\section{Visualization}
\label{sec:apH}
We visualize the model’s node-graph hierarchcal attention scores and prediction results, as shown in the Figure \ref{pic:result_example.pdf}. In this example, the historical event chain consists of events \textit{[open, a man in a red plaid shirt, a door]}, \textit{[reach, woman in a black coat, a drumset]}, \textit{[smile, woman in a black coat, man in a white shirt]}, \textit{[play, a man in a red plaid shirt, a music, a guitar]}, while the ground-truth future event is \textit{[play, woman in a black coat, a music, a drumset]}. The model, however, predicted \textit{[play, man in a white shirt, a music, a guitar]}. At the graph-level attention, the third and fourth events received the highest attention scores $0.417$ and $0.491$, which aligns well with our intuition. At the node level, the verb node \textit{play} and the noun node \textit{man in a white shirt} received the highest attention scores $0.458$ and $0.154$. This contributed to the model correctly predicting the future event’s verb, but incorrectly predicting the associated noun.

\end{document}